\documentclass[11pt]{article}
\usepackage[margin=1in]{geometry}
\usepackage[T1]{fontenc}
\usepackage[utf8]{inputenc}
\usepackage{lmodern}
\usepackage[protrusion=true,expansion=false]{microtype}
\usepackage{graphicx}
\usepackage{booktabs}
\usepackage[dvipsnames]{xcolor}
\usepackage{amsmath}
\usepackage{amssymb}
\usepackage{mathtools}
\usepackage{amsthm}
\usepackage{bm}
\usepackage{comment}
\usepackage{natbib}
\setcitestyle{authoryear}
\usepackage{pgfplots}
\usepackage[scientific-notation=true]{siunitx}
\pgfplotsset{compat=1.9}
\usepackage[ruled]{algorithm2e}

\SetAlFnt{\small}
\SetAlCapFnt{\small}
\SetAlCapNameFnt{\small}
\SetAlCapHSkip{0pt}
\usepackage{algorithmic}
\usepackage{tikz}
\usetikzlibrary{fit}
\usetikzlibrary{positioning}
\usepackage{multirow}
\usepackage{array}
\usepackage{enumitem}
\usepackage{tabularx}
\usepackage{hhline}
\usepackage{nicefrac}
\usepackage{caption}
\usepackage{subcaption}
\usepackage{placeins}
\usepackage{url}
\usepackage{hyperref}
\hypersetup{pdftitle={Proper Dataset Valuation by Pointwise Mutual Information}, pdfauthor={Rui Ray Chen, Xuan Qi, Yuxin Chen, Yongchan Kwon, James Zou, Shuran Zheng}}
\usepackage[capitalize,noabbrev]{cleveref}

\newcommand{\X}[0]{\mathbf{X}}
\newcommand{\x}[0]{{\mathbf{x}}}
\newcommand{\w}[0]{{\mathbf{w}}}
\newcommand{\z}[0]{{\mathbf{z}}}
\newcommand{\bth}[0]{{\bm{\theta}}}
\newcommand{\y}[0]{{\mathbf{y}}}

\renewcommand{\hat}{\widehat}
\renewcommand{\tilde}{\widetilde}
\newcommand{\E}[0]{\mathbb{E}}

\newcommand{\Description}[1]{}

\definecolor{accessblue}{rgb}{0.0, 0.28, 0.67}
\definecolor{accessorange}{rgb}{0.90, 0.45, 0.0}

\theoremstyle{plain}
\newtheorem{theorem}{Theorem}[section]
\newtheorem{proposition}[theorem]{Proposition}
\newtheorem{lemma}[theorem]{Lemma}
\newtheorem{corollary}[theorem]{Corollary}

\theoremstyle{definition}
\newtheorem{definition}[theorem]{Definition}

\theoremstyle{remark}

\title{Proper Dataset Valuation by Pointwise Mutual Information}

\author{
\small
\begin{tabular}{cc}
Rui Ray Chen\thanks{Both authors contributed equally to this research.} & Xuan Qi\footnotemark[1] \\
Tsinghua University & Tsinghua University \\
\texttt{chenrui20@mails.tsinghua.edu.cn} & \texttt{qi-x22@mails.tsinghua.edu.cn} \\
\addlinespace[0.8em]
Yuxin Chen & Yongchan Kwon \\
Tsinghua University & Together AI \\
\texttt{chenyuxin25@mails.tsinghua.edu.cn} & \texttt{ykwon@together.ai} \\
\addlinespace[0.8em]
James Zou & Shuran Zheng \\
Stanford University & Tsinghua University \\
\texttt{jamesz@stanford.edu} & \texttt{shuran0zheng@gmail.com}
\end{tabular}
}
\date{}

\begin{document}

\maketitle

\begin{abstract}
Data plays a central role in advancements in modern artificial intelligence, with high-quality data emerging as a key driver of model performance.
This has prompted the development of principled and effective data curation methods in recent years. However, existing methods largely rely on heuristics, and whether they are truly effective remains unclear.
For instance, standard evaluation methods that assess a trained model's performance on specific benchmarks may incentivize assigning high scores to data that merely resembles the test set.
This issue exemplifies Goodhart's law: when a measure becomes a target, it ceases to be a good measure.
To address this issue, we propose an information-theoretic framework for evaluating data curation methods. We define dataset quality in terms of its informativeness about the true model parameters, formalized using the \emph{Blackwell ordering of informativeness}. Under this ordering, Blackwell's theorem ensures that more informative data yields optimal models with lower expected loss on the true underlying distribution. To measure informativeness, we show that the Blackwell order can be determined by the Shannon mutual information between the curated data and the test data. To estimate this mutual information, we introduce a novel method that trains Bayesian models on embedded datasets and computes mutual information from the posteriors of model parameters.
Experiments on real-world data demonstrate that our mutual information-based evaluation assigns appropriately lower scores to data curation strategies that reduce dataset informativeness, while traditional test score-based evaluation methods may favor data curation strategies that overfit to the test set but compromise the training data's informativeness.
\end{abstract}

\newpage

\section{Introduction}

Data plays a central role in the development of modern artificial intelligence (AI) systems, where both the volume and quality of training data are critical to model performance
\citep{brown2020language, peebles2023scalable, team2024gemma, grattafiori2024llama3herdmodels}. As AI systems continue to scale and the computational costs of training escalate, the focus is shifting from simply expanding model and dataset sizes to enhancing the quality of the data itself. This shift has prompted the development of various data curation strategies, including data filtering \citep{gunasekar2023textbooks, li2023textbooks, fang2023data, pouget2024no, li2025datacomplmsearchgenerationtraining}, duplicate removal \citep{kandpal2022deduplicating, abbas2023semdedupdataefficientlearningwebscale}, data augmentation \citep{muennighoff2024scaling}, and synthetic data generation \citep{liu2024best}.

However, ensuring the effectiveness of these curation techniques remains a major challenge \citep{li2025datacomplmsearchgenerationtraining, weber2024redpajama}. The standard evaluation approach involves training a model on a curated dataset and measuring its performance against benchmark test sets \citep{li2025datacomplmsearchgenerationtraining,albalak2024surveydataselectionlanguage, team2024gemma, grattafiori2024llama3herdmodels}. This methodology, though common, can inadvertently incentivize undesirable curation practices that optimize performance on specific benchmarks, yet risk overfitting to test data and undermining the model's generalization capabilities. Recently, \citet{mizrahi2025languagemodelsimprovepretraining} demonstrated that selecting pretraining data to match specific benchmarks (``CORE'' benchmarks) boosts test scores on these benchmarks but significantly degrades performance on other tasks (``NON-CORE'' benchmarks).
This performance disparity illustrates a critical issue: as highlighted by Goodhart's law, when a measure becomes a target, it ceases to be a good measure.

A critical question, therefore, is how to distinguish data curation methods that merely manipulate data to boost model performance on specific benchmarks from those that truly contribute meaningful data to the learning process.
In this work, we propose an alternative information-theoretic framework that may help make this distinction: rather than measuring the test score of a trained model on specific test sets, we evaluate the \emph{informativeness} of a dataset for a given machine learning task. To achieve this, we adopt the well-known Blackwell ordering~\citep{Experiments:112749} to compare the informativeness of datasets. Under this ordering, Blackwell’s theorem ensures that more informative data yields optimal models with lower expected loss on the true underlying distribution. Therefore, a data curation method is considered effective if it increases the dataset's informativeness about the true model parameters, while it is deemed \emph{strategic} if it decreases the dataset's informativeness, according to the Blackwell ordering.

To quantify informativeness, we propose using the \emph{Shannon mutual information} (MI) of the curated dataset and the test dataset as a metric. We show that MI effectively identifies data curation methods that truly reduce informativeness. Specifically, if a curation method decreases the dataset's informativeness according to the Blackwell ordering, it must lead to a decrease in the mutual information.
However, estimating the mutual information of two \emph{datasets} introduces a new challenge. Prior work on high-dimensional mutual information estimation has primarily focused on pairs of individual data points with dimensionality on the order of thousands (e.g., correlated image pairs from MNIST or CIFAR-10), but not on dataset pairs containing many such points.\footnote{\citet{gowri2024approximating} show that existing neural estimators typically handle only tens of \emph{intrinsic dimensions}.
This is especially problematic, as datasets are inherently high-dimensional due to the large number of data points they contain. Given that the intrinsic dimensionality of images ranges from 20 to 43, as reported by \citet{pope2021intrinsic}, existing methods may not be reliable when applied to datasets with just a few images.}
To overcome this challenge, we propose a novel method for estimating the mutual information of two datasets. We exploit a dataset's capacity to train a machine learning model and compute the mutual information using the posterior distributions of model parameters obtained from Bayesian models.

We demonstrate the effectiveness of our method through experiments on real-world datasets, ranging from standard image datasets (MNIST, CIFAR, ImageNet) to large-scale language preference tasks (SHP, Skywork). First, we show that for high-dimensional dataset pairs, our MI estimator yields much more accurate estimates compared to previous baseline methods like MINE \citep{belghazi2021minemutualinformationneural} and LMI \citep{gowri2024approximating}, with theoretical support from a faster convergence rate that is independent of input dimensionality. Second, we show that the test score-based evaluation method may favor data curation strategies that make the dataset more similar to the test data but reduce its informativeness about the true model parameters. In contrast, our mutual information-based evaluation assigns appropriately lower scores to such strategies. This is because our method accurately estimates the mutual information of datasets, as verified by our experiments. Finally, we validate this advantage in realistic OOD scenarios. We demonstrate that while standard accuracy metrics can be inversely correlated with robustness—misleadingly favoring redundant data—our PMI score maintains a strict alignment with true out-of-distribution generalization. This confirms that our metric effectively serves as a reliable proxy for genuine parameter learning, distinguishing high-quality diverse data from those that merely overfit the test distribution.

\section{Related Work} \label{sec:related-work}

\paragraph{Data curation.} Due to the growing importance of data quality, recent years have seen a surge in diverse data curation techniques.
Data filtering \citep{gunasekar2023textbooks, li2023textbooks, fang2023data, pouget2024no, xie2023dataselectionlanguagemodels, li2025datacomplmsearchgenerationtraining} focuses on enhancing the intrinsic quality of the dataset by selecting high-value data points and discarding irrelevant or low-quality content based on various heuristics.
Complementarily, duplicate removal \citep{kandpal2022deduplicating, abbas2023semdedupdataefficientlearningwebscale} addresses repeated occurrences and the impact of sequences within training datasets. The findings underscore the importance of sequence-level deduplication in training efficiency and model privacy without sacrificing model performance. Data augmentation \citep{muennighoff2024scaling} generates new training samples from the original dataset to enhance its diversity and volume while preserving its core characteristics, while synthetic data generation \citep{liu2024best} creates new data that closely resemble the distribution of real data. Data mixing \citep{xie2024doremi, liu2025regmixdatamixtureregression} determines the weight of each domain’s dataset to optimize performance across all domains. Data distillation \citep{sachdeva2023datadistillationsurvey} aims to create compact, high-fidelity data summaries that capture the most essential knowledge from a given target dataset. MeCo \citep{DBLP:journals/corr/abs-2501-01956} optimizes the data curation process by injecting additional metadata into the data, further refining its utility in model training. \citet{DBLP:journals/corr/abs-2502-10341} introduce the WebOrganizer framework, which classifies data based on both topics and formats before applying mixing optimization, thereby effectively enhancing the quality of pretraining data. For a more comprehensive survey of related work in this area, we refer readers to \citep{DBLP:conf/icml/WettigGM024}.

\paragraph{Dataset valuation.} Various methods have been proposed for dataset evaluation. The standard approach involves training a model on a curated dataset and measuring its performance on benchmark test sets \citep{li2025datacomplmsearchgenerationtraining,albalak2024surveydataselectionlanguage}. However, this empirical approach is highly susceptible to dataset shift and benchmark unreliability; for instance, \citet{pmlr-v97-recht19a} demonstrate that model accuracy can drop substantially on newly constructed test sets despite using matched collection procedures, highlighting that benchmark performance may not fully capture robust generalization. To provide more intrinsic assessments, \citet{garridolucero2024dushapleyshapleyvalueproxy} leverages estimated Shapley values for efficient dataset valuation. \citet{MohammadiAmiri_Berdoz_Raskar_2023} focus on intrinsic, task-agnostic dataset valuation by estimating data diversity and relevance without requiring a validation set. However, none of these methods provide the information-theoretic guarantees as we do. \citet{pmlr-v119-sim20a} evaluate the contribution of datasets by the mutual information of the model parameter and the data; however, their method does not detect strategic data curation methods. See \Cref{app:scoring_rules} for related discussion.

\paragraph{Data point valuation.} Besides dataset evaluation, the assessment of data point value has been actively studied in the data valuation literature. A standard approach is to measure the change in the test accuracy after removing a single training data point of interest. Data Shapley by \citet{ghorbani2019data} deploy the Shapley value from cooperative game theory to ML settings, and several variants that improve its computational efficiency or relax underlying conditions have been proposed \citep{jia2019efficient, kwon2021beta, wang2022data, wang2024data, pham2025dupredatautilityprediction, wang2025data, wang2024efficientdatashapleyweighted}. An alternative common approach utilizes the influence function introduced in robust statistics \citep{koh2017understanding, feldman2020neural}. This method provides a mathematically rigorous interpretation of data values and has been implemented in various applications, such as image classification and sentiment analysis, or text-to-image generation \citep{park2023trak, kwon2023datainf}. Another related line is datamodels, which learn a mapping from training-set membership to model predictions and use it to study how training examples affect model behavior~\citep{pmlr-v162-ilyas22a}. \citet{wang2024capturingtemporaldependencetraining} efficiently capture temporal data influence in training while \citet{wang2024greats} optimize LLM training via Taylor-based batch selection. Other algorithm-agnostic and task-agnostic methods have also been explored, such as \citep{just2023lavadatavaluationprespecified,NEURIPS2021_59a3adea, lin2025efficient}. There are also some theoretical works, \citet{wang2024rethinkingdatashapleydata} critically analyze Data Shapley’s limitations in selection tasks while \citet{pmlr-v235-lin24t} introduce model deviation and NTK theory for validation-free robustness.
We refer the readers to \citet{jiang2023opendataval} for a comprehensive and detailed review.

\paragraph{Peer prediction approach.} Our method is also connected to the \emph{peer prediction} literature \citep{MRZ05,prelec2004bayesian,jurca2008incentives,radanovic2013robust,radanovic2014incentives,witkowski2012peer,kong2016equilibrium,schoenebeck2020two}, which studies eliciting truthful information without ground truth. Among these, \citet{kong2018water,chen2020truthful,schoenebeck2020learning} are most relevant. \citet{kong2018water} proposed a mutual information-based peer prediction method using two agents' predictions about a latent label, later adapted for data valuation by \citet{chen2020truthful}. However, their approach computes pointwise mutual information through a complex integral involving the product of two posteriors divided by the prior (see \Cref{app:integral} for details). \citet{burrell2022measurementintegritypeerprediction} introduce the notion of measurement integrity in peer prediction, studying whether peer-prediction mechanisms preserve meaningful comparisons of report quality. \citet{schoenebeck2020learning} also estimates mutual information but is restricted to discrete variables or specific continuous distributions. Recently, \citet{lu2024elicitinginformativetextevaluations} adapted peer prediction principles for reference-free LLM evaluation, proposing a framework grounded in Blackwell dominance that scores candidate responses based on their pointwise mutual information with independent high-quality judgments. Similarly, \citet{xu2025benchmarkingllmsjudgmentsgold} extends classic peer prediction mechanisms to high-dimensional text signals, utilizing LLM-approximated posteriors and logarithmic scoring rules to incentivize high-effort text generation while employing synopsis conditioning to filter out easily replicable shortcut information.

\paragraph{Mutual information estimation.}
Mutual information (MI) is a key concept in data science that measures the statistical dependence between random variables. However, estimating mutual information $I(Y, Z)$ given $k$ samples $(Y_1, Z_1), \dots, (Y_k, Z_k) \sim p(Y, Z)$ in high-dimensional settings is notoriously difficult. Non-parametric methods, such as binning, likelihood-ratio estimators with support vector machines, and kernel-density estimators, are commonly used \citep{fraser1986independent, darbellay1999estimation, kraskov2004estimating} for estimating mutual information, but these approaches often do not scale well with sample size and data dimensionality \citep{gao2015efficient}. Variational neural estimators, such as MINE \citep{belghazi2021minemutualinformationneural} and InfoNCE \citep{oord2018representation}, have become popular alternatives.  These neural MI estimators fall into two main categories: \emph{discriminative} and \emph{generative} approaches~\citep{song2019understanding}.
    Discriminative approaches (e.g.,~\citep{belghazi2021minemutualinformationneural, oord2018representation, liao2020demi}) estimate MI by approximating the density ratio directly from samples. However, these methods suffer from the \emph{high-discrepancy issue}, where the variance of the estimator can grow exponentially in high-dimensional settings~\citep{pmlr-v108-mcallester20a, song2019understanding}.
     Generative approaches (e.g.,~\citep{butakov2024mutual, franzese2023minde}) aim to estimate the distributions from samples using generative models. The accuracy of these estimates heavily depends on the quality of the learned generative models,  and generative modeling is well-known hard in high-dimensional cases \citep{chen2020neural,lipman2022flow}.
Prior work has primarily focused on pairs of individual data points, typically in the range of thousands of dimensions  (e.g., correlated image pairs from MNIST or CIFAR-10).
Recent work by \citet{gowri2024approximating} shows that while standard MI estimators perform well in up to tens of intrinsic dimensions, they are not reliable when intrinsic dimensionality gets higher and available data is limited. To mitigate this, they suggest reducing dimensionality with pre-trained models before MI estimation, which improves scalability.

\section{Model} \label{sec:model}

Consider a machine learning task with a model parameterized by \( \bth \in \Theta \subseteq \mathbb{R}^k \). We assume the Bayesian framework, where a parameter \( \bth \) is drawn from an underlying prior distribution \( p(\bth) \). Suppose we have a test dataset \( T = (\x_T^{(1)}, \dots, \x_T^{(N_T)}) \), consisting of \( N_T \) independent and identically distributed (i.i.d.) data points drawn from an underlying distribution \( p(\x_T|\bth) \), and an original dataset \( D = (\x_D^{(1)}, \dots, \x_D^{(N_D)}) \), with \( N_D \) i.i.d. data points from an underlying \( p(\x_D|\bth) \). The two datasets may not follow the same distribution, so \( p(\x_D|\bth) \) need not equal \( p(\x_T|\bth) \). Denote the support of \( D \) and \( T \) by \( \mathcal{D} \) and \( \mathcal{T} \), respectively.

We aim to evaluate different data curation methods, which can be seen as functions applied to the original dataset, possibly incorporating additional information to improve data quality.
This additional information is represented by a random variable \( A \), which may be correlated with both the model parameter \( \bth \) and the dataset \( D \). We formalize this concept below.

\begin{definition}[Data curation method]
    Let \( A \in  \mathcal{A}\) be a random variable representing additional information for data curation. A data curation method with additional information \( A \) is a function \( f: \mathcal{A} \times \mathcal{D} \to \mathcal{D} \) that outputs a modified dataset \( f(A, D) \) given \( A \) and an original dataset \( D \in \mathcal{D} \). The space of such functions is denoted by \( \mathcal{F} \).
\end{definition}

Below are several examples of data curation methods:
\begin{itemize}

	\item \textbf{Adding new data.} A simple data curation method is adding new data, where $A\in \mathcal{A} = \mathcal{D}$ represents the new data and $f(A,D) = D\cup A$.

	\item \textbf{Deleting data.} It is also common to select a subset of data and remove the others, as seen in coreset selection~\citep{mirzasoleiman2020coresets}, data filtering with quality signals~\citep{gunasekar2023textbooks, li2023textbooks, fang2023data, pouget2024no, li2025datacomplmsearchgenerationtraining}, data deduplication~\citep{kandpal2022deduplicating, abbas2023semdedupdataefficientlearningwebscale}, and removing low-quality or out-of-domain data~\citep{northcutt2021confident, ghorbani2019data, jia2019towards}. In data deletion, the additional information can be represented as  a random vector \( A \in \{0, 1\}^{N_D} \) indicating whether each data point is retained or removed.

	\item \textbf{Reweighting data.}
	Another commonly used method is resampling data points with different weights~\citep{xie2024doremi,xu2024bayesianapproachdatapoint}. Here the additional information can be represented by a random vector \( A \in \mathbb{N}^{N_D} \) indicating the number of copies of each data point in the final dataset.

\end{itemize}

To rigorously distinguish between methods that merely adapt the dataset to be more similar to the test data and those that introduce meaningful improvements, we employ the \emph{Blackwell ordering of informativeness}.
Rather than relying on empirical test accuracy---which can be confounded by biases in a specific test set $T$---we evaluate data curation methods by their effect on the \emph{informativeness} of the dataset $D$ about the true underlying model parameters $\theta$.

\begin{definition}[Blackwell order of informativeness~\citep{Experiments:112749}]
    For random variables $X$, $Y$, and $Z$, we say \( Z \) is \emph{less informative} than \( Y \) about \( X \) if these random variables form a Markov chain \( X \to Y \to Z \).
\end{definition}

The introduction of Blackwell ordering is justified by the following theorem.
Consider a method \( f(A, D) \) that reduces the
informativeness of \( D \) about the true model parameter \( \bth \) according
to the Blackwell order. Then, by Blackwell's theorem, the optimal model trained
on \( D \) can achieve an expected loss (over the true underlying distribution) that is no greater than that achievable
by the optimal model trained on \( f(A, D) \).

\begin{theorem}[Informal, \citep{Experiments:112749}]
    Suppose \( \bth \to D \to f(A,D) \) forms a Markov chain. Consider the decision problem of selecting a model \( h \) from a model class \( \mathcal{H} \) to minimize the expected loss using a dataset. Then, the minimum expected loss achievable using \( D \) is at least as low as that achievable using \( f(A,D) \), where the expected loss is defined over the true underlying distribution by any training loss function.
    The formal version can be found in~\Cref{app:blackwell}.
\end{theorem}

We therefore define \emph{strategic data curation methods} as those that reduce the dataset's informativeness about the true model parameter \(\bth\), even if they may improve test scores.

\begin{definition}[Strategic data curation] \label{def:strategic}
   Consider the joint distribution $p(\bth,D,A)$. A data curation method \( f(\cdot) \) is strategic if the curated dataset \( f(A, D) \) is less informative about \( \bth \) than the original dataset \( D \). Formally, \( \bth \to D \to f(A, D) \) forms a Markov chain.
\end{definition}

Below are several examples of strategic curation methods:
\begin{itemize}

	\item \textbf{Adding recursively generated synthetic data.} When $f(A,D)$ adds new data $A$ to $D$, if \( A \) consists of synthetic data produced by a model $\hat \bth$ trained on \( D \), then \( f(A, D) \) is strategic because \( \bth \to D \to \hat{\bth} \to A \) forms a Markov chain. As highlighted by \citet{Shumailov2024AI}, the use of such recursively-generated data may cause \textit{model collapse}. In contrast, if \( A \) contains synthetic data generated from unseen sources or filtered by verifiers, it may introduce new information, making \( f(A, D) \) non-strategic.

	\item \textbf{Deleting or reweighting data without additional signals.} When deleting or reweighting data, if \( A \in \mathbb{N}^{N_D} \) is guided by some additional quality or relevance signal, the filtered/reweighted dataset can be more informative, making $f(A,D)$ non-strategic. Conversely, if \( A \) is decided solely from the observed dataset $D$ without utilizing new signals,
		the resulting dataset $f(A,D)$ will be less informative. Because
		$\bth \to D \to A$ forms a Markov chain, and thus $\bth \to D \to f(A,D)$ forms a Markov chain, making $f(A,D)$ strategic.

	\item \textbf{Deleting or reweighting data by non-essential features.} In addition, when deleting or reweighting data, if \( A \in \mathbb{N}^{N_D} \) (the number of copies of each data point in the final dataset) is based on some non-essential feature that is non-predictive of the label, the resulting dataset will be less informative, making $f(A,D)$ strategic.
	To be more specific, suppose a data point $\x = (\z,y)$ in $D$ consists of a label $y$ and essential features $\z$. Suppose there is some non-essential feature $z_N$ that satisfies $p(y|\bth, \z, z_N)=p(y|\bth, \z)$ and is non-predictive of $y$ conditioned on $\z$, as illustrated in the graphical model in \Cref{fig:example_GM}.
    Then if the vector \( A \in \mathbb{N}^{N_D} \) is decided by this non-essential feature of the data points, $z_N^{(1)}, \dots, z_N^{(N_D)}$ (as well as $D$), then the resulting dataset will be less informative because $z_N^{(1)}, \dots, z_N^{(N_D)}$ are independent of $\bth$ conditioned on $D$ (as the path between $\bth$ and $z_N$ is blocked by $\z$ and d-separation implies conditional independence).

\end{itemize}

\begin{figure}[t]
    \centering

    \begin{tikzpicture}[
    node distance=0.5cm and 0.5cm,
    every node/.style={draw, circle},
    thick,
    plate/.style={draw, rectangle, fit=#1}
    ]
    \node (ty) {$\bth$};
    \node (Y) [right=of ty] {$y$};
    \node (C) [right=of Y] {$\z$};
    \node (X) [right=of C]{$\z, z_N$};
    \node (S) [right=of X] {$z_N$};

    \draw [->] (X) -- (C);
    \draw [->] (X) -- (S);
    \draw [->] (C) -- (Y);
    \draw [->] (ty) -- (Y);

    \node[plate=(Y)(C)(X)(S)] (plateYZ) {};
    \end{tikzpicture}
    \caption{Graphical model for non-essential features. The model parameter \(\theta\) and essential feature \(\mathbf z\) determine the label \(y\), while the observed input contains both \(\mathbf z\) and a non-essential attribute \(z_N\). Since \(z_N\) is non-predictive of \(y\) conditional on \(\mathbf z\), curation based only on \(z_N\) can change dataset composition without adding information about \(\theta\).}
    \Description{A directed graphical model showing that theta and the essential feature determine the label, while the observed input contains both the essential feature and a non-essential attribute.}
    \label{fig:example_GM}
\end{figure}

The remaining question is how to determine whether $\theta$, $D$, and $f(A, D)$ form a Markov chain. Rather than addressing this relation directly, we consider designing a \emph{scoring function} that quantifies the informativeness of curated datasets.
Specifically, we evaluate data curation methods by assigning a score that reflects their impact on informativeness, with the goal of distinguishing between methods that increase or reduce informativeness.

\begin{definition}[Scoring function for data curation methods]
    A scoring function for data curation methods \( S: \mathcal{F}\times \Delta(\mathcal{D}\times \mathcal{T}) \to \mathbb{R} \) assigns a score \( S(f(\cdot), D, T) \) to a data curation method \( f(\cdot) \),\footnote{The notation $\Delta(\mathcal{D}\times \mathcal{T})$ represents the space of all distributions over $\mathcal{D}\times \mathcal{T}$. This means that we allow the scoring function to sample datasets from the underlying distribution and determine the score via the samples.} given access to the original data \( D \) and test data \( T \). For simplicity, we omit $D$ and $T$ and write $S(f(\cdot))$ when clear from the context.
\end{definition}

In this work, we aim to design a scoring function that does not encourage strategic data curation methods. Specifically, we seek a function that assigns lower scores to strategic methods than to the case of no modification.

\begin{definition}[Strategy-proof scoring functions]
    A scoring function $S(f(\cdot))$ is \emph{strategy-proof} if:
    \begin{itemize}
        \item For any strategic method $f(\cdot)$, we have $S(f(\cdot)) \leq S(\mathrm{id}(\cdot))$, where $\mathrm{id}(A,D) \equiv D$;
        \item $S(\cdot)$ is non-constant (i.e., not all methods receive the same score).
    \end{itemize}
\end{definition}

Given \( D \), \( T \), and a data curation method \( f(\cdot) \), a natural scoring function is to train a model on curated data \( f(A,D) \) and use its test accuracy on \( T \) as the score. However, as we demonstrate in~\Cref{sec:exp_scoring_evaluation}, this approach is not strategy-proof: it may incentivize curation strategies that make \( D \) more similar to \( T \),  improving test accuracy while reducing the informativeness of the  dataset.

\section{PMI Dataset Score} \label{sec:framework}

Designing a strategy-proof scoring function is not straightforward. We propose using the  \emph{Shannon mutual information} (MI) of the curated data $f(D)$ and the test data $T$, which can be proved to constitute a strategy-proof scoring function.
However, estimating MI in high-dimensional settings is notoriously challenging. \citet{gowri2024approximating} show that existing neural estimators typically handle only tens of \emph{intrinsic dimensions}.
This is especially problematic, as datasets are inherently high-dimensional due to the large number of data points they contain. Given that the intrinsic dimensionality of images ranges from 20 to 43, as reported by \citet{pope2021intrinsic}, existing methods may not be reliable when applied to datasets with just a few images or text documents. To overcome this challenge, we leverage a key property of datasets:
\begin{center}
\fbox{
\begin{minipage}{0.9\textwidth}
\centering
\itshape A dataset consists of i.i.d. data points and can be used to train a machine learning model.
\end{minipage}
}
\end{center}
By utilizing this property, we introduce a novel method for approximating the \emph{pointwise mutual information} (PMI) of datasets. This enables accurate and efficient estimation built on established Bayesian machine learning methods.
In this section, we omit $A$ and use $f(D)$ to represent a data curation method for simplicity.

\subsection{Mutual information as the metric}
We first propose using the \emph{Shannon mutual information} of the curated dataset $\widehat{D} = f(D)$ and the observable test dataset $T$ as our scoring function, which serves as a strategy-proof scoring function.\footnote{Another metric that might yield a strategy-proof scoring function is the Shannon mutual information between the model parameters $\bth$ and the curated dataset $f(D)$, denoted $I(\bth, f(D))$.
However, this metric does not prevent strategic data curation methods unless the underlying true model parameter $\bth$ is observable. We defer the detailed discussion to~\Cref{app:scoring_rules}.}

\begin{proposition} \label{prop:MI_metric}
Given a data curation method $f(\cdot)$, let $\widehat{D} = f(D)$. Then the \emph{Shannon mutual information} $I(\widehat{D}, T)$, when computable, constitutes a strategy-proof scoring function. Formally,
\[
I(\widehat{D}; T) = \mathbb{E}_{\widehat{D},T}\left[PMI(\widehat{D}, T)\right] := \mathbb{E}_{\widehat{D},T}\left[\log \frac{p(\widehat{D},T)}{p(\widehat{D})p(T)}\right]
\]
where:
\begin{itemize}[leftmargin=*, itemsep=0pt, topsep=0pt]
\item $PMI(\widehat{D}, T) = \log \frac{p(\widehat{D},T)}{p(\widehat{D})p(T)}$ denotes the pointwise mutual information.
\item The expectation is taken over the joint distribution $p(\widehat{D}, T)$ induced by $f(\cdot)$ and the data generating process in \Cref{sec:model}, with \( T = (\x_T^{(1)}, \dots, \x_T^{(N_T)}) \) and \( D = (\x_D^{(1)}, \dots, \x_D^{(N_D)}) \) being datasets of multiple data points.
\end{itemize}
The proof is deferred to \Cref{app:MI_metric}.
\end{proposition}

\paragraph{Main challenge: high-dimensional MI estimation is difficult.} Then the problem boils down to estimating the mutual information of two \textbf{datasets} $\widehat D,T$ that contain \textbf{many data points}, where the total dimensionality scales as \(\bm{\textbf{dim}(T) = \textbf{dim}(\mathbf{x}_T) \cdot N_T,}\) with $N_T$ being the number of data points and $\text{dim}(\mathbf{x}_T)$ the dimensionality of each individual data point.
However, estimating MI in high-dimensional settings is notoriously difficult. Each existing method (e.g.~\citep{belghazi2021minemutualinformationneural, oord2018representation, liao2020demi, butakov2024mutual, franzese2023minde}) has significant limitations. A more detailed discussion of these methods can be found in \Cref{sec:related-work}.
In particular, prior work has primarily focused on pairs of individual data points with dimensionality on the order of thousands (e.g., correlated image pairs from MNIST or CIFAR-10)---a challenging task in itself.
Our setting presents a more severe challenge: the aggregate dimensionality of the full datasets becomes prohibitively large due to the number of data points.

\subsection{Closed-form approximation of pointwise mutual information}

To overcome this challenge, we propose a novel mutual information estimation method that leverages a dataset's ability to train a machine learning model. A key contribution is a novel closed-form expression for pointwise mutual information (PMI), enabling accurate and efficient estimation built on established Bayesian machine learning methods~\citep{jospin2022hands}.

Our method begins by generating $k$ dataset pairs $(D_1, T_1), \dots, (D_k, T_k)$ and estimating mutual information $I(f(D), T)$ via the average:
\(
\frac{1}{k} \sum_{i=1}^k \text{PMI}(f(D_i), T_i)
\)
defined in~\Cref{prop:MI_metric}.
To approximate PMI, we
train Bayesian models (such as Bayesian logistic regression outlined in \Cref{app:example} and Bayesian neural networks) on these datasets. Let the Bayesian model have parameters $\theta \in \Theta$, prior $p(\theta)$, and likelihoods $p(\x_D|\theta)$ and $p(\x_T|\theta)$ for the training and the test data.
Finally, we compute PMI by a novel closed-form formula, assuming that $p(\theta|\cdot)$ can be approximated by tractable distributions via established posterior inference techniques for Bayesian models.\footnote{Extensive research in Bayesian machine learning has focused on estimating the posterior distribution of model parameters $p(\theta|\cdot)$. For instance, Laplace approximation and variational inference approximate the posterior $p(\theta|\cdot)$ by tractable distributions. But even with a tractable approximate posterior \( p(\theta|\cdot) \), the \emph{posterior predictive} \( p(T_i|f(D_i)) = \int_\theta p(T_i|\theta)p(\theta|f(D_i)) \,d\theta \) in PMI remains intractable for most models, including logistic regression. Common approximation approaches each have limitations in our setting: (1) Monte Carlo integration suffers from numerical instability with near-zero likelihoods $p(T_i|\theta)$; probit approximation~\citep{gibbs1998bayesian} and Laplace bridge methods~\citep{hobbhahn2022fast} are limited to single-point predictions \( \int_\theta p(x_T^{(i)}|\theta)p(\theta|D) \,d\theta \) whereas we require  joint predictions  \( \int_\theta \prod_i p(x_T^{(i)}|\theta)p(\theta|D) \,d\theta, \)  for entire datasets.
	}

\begin{theorem}[PMI dataset score]
	 \label{thm:MI}
Let $f(\cdot)$ be a data curation method producing curated datasets $\widehat{D}_i = f(D_i)$. Suppose we have a Bayesian model over the datasets with parameters $\theta \in \Theta$, prior $p(\theta)$, and likelihoods $p(\x_D|\theta)$ and $p(\x_T|\theta)$. Let $p(\theta|X)$ denote the posterior distribution of $\theta$ given dataset $X$. Then $PMI(\widehat D_i, T_i)$ can be computed as
\begin{align} \label{eqn:MI_score}
  PMI(\widehat D_i, T_i)= & U_\eta(\widehat D_i, T_i) :=  \log  \frac{p(\theta =\eta|\widehat D_i)\cdot p(\theta = \eta|T_i)}{p(\theta = \eta) \cdot p(\theta = \eta|\widehat D_i, T_i)},
\end{align}
where $\eta$ is an arbitrary parameter value in $\Theta$.
\end{theorem}

To prove the theorem, we first prove the following lemma.
\begin{lemma}\label{lem:likelihood_posterior}
Let $D$ and $T$ be two random variables that are independent conditional on random variable $\bth$, that is, $p(D,T|\bth) = p(D|\bth)p(T|\bth)$. Then we have for any $\eta\in \Theta$, $d \in \mathcal{D}$, and $t\in\mathcal{T}$,
\begin{align*}
    \frac{p(T = t|D=d)}{p(T = t)} = \frac{p(\bth =\eta|D = d)\cdot p(\bth = \eta|T = t)}{p(\bth = \eta) \cdot p(\bth = \eta|D=d, T=t)}.
\end{align*}
\end{lemma}
The proof of~\Cref{lem:likelihood_posterior} mainly relies on Bayes' rule and the conditional independence condition.
\begin{proof}
Since $D,T$ are independent conditional on $\bth$, for any $\eta \in \Theta$ we have
\begin{align*}
	&p(\bth = \eta|D=d, T=t) \\
	& = \frac{p(D=d, T=t|\bth = \eta)\cdot p(\bth = \eta)}{p(D=d, T=t)}\\
	& = \frac{p(D=d|\bth = \eta)\cdot p(T=t|\bth = \eta)\cdot p(\bth = \eta)}{p(D=d, T=t)}\\
	& = \frac{p(\bth = \eta|D=d)\cdot p(\bth = \eta|T=t) \cdot p(D =d) \cdot p(T=t)}{p(\bth = \eta)\cdot p(D=d, T=t)}.
\end{align*}
Then we have
	\begin{align*}
		 \frac{p(\bth =\eta|D = d)\cdot p(\bth = \eta|T = t)}{p(\bth = \eta) \cdot p(\bth = \eta|D=d, T=t)} 	&= \frac{p(D=d, T=t)}{p(D =d) \cdot p(T=t)}\\
	& =\frac{p(T = t|D=d)}{p(T = t)}.
	\end{align*}
\end{proof}
 With this equation, we can apply the logarithmic scoring rule to get a truthful valuation function, which gives the valuation function in~\Cref{thm:MI}. The proof is as follows.
\begin{proof}
According to~\Cref{lem:likelihood_posterior}, $U(d,t) = \log p(T=t|D=d)/P(T=t)$. Then the expected score is maximized by reporting $d$ because
\begin{align*}
& \E_{T} [U_\eta(d, T)|D = d] - \E_{T} [U_\eta(d', T)|D = d]	\\
= & \int_t p(t|D=d) \log p(t|D=d) \, dt - \int_t p(t|D=d) \log p(t|D=d') \, dt\\
= & \int_{t} p(t|D=d) \log \frac{p(t|D=d)}{p(t|D=d')} \, dt\\
= & \ D_{KL}\big(p(t|D=d), p(t|D=d')\big) \\
\ge & \ 0.
\end{align*}
And when truthful reporting, the expected score $\E[U_\eta(D,T)]$ is just the Shannon mutual information $I(D,T) = \E_{D,T}\left[\log \frac{p(D,T)}{p(D)p(T)}\right]$.
\end{proof}

\paragraph{Key advantages.} A key advantage of our proposed PMI dataset score is its modularity.
The streamlined form of the PMI dataset score allows for broad integration with various Bayesian machine learning frameworks. Specifically, it is compatible with Bayesian neural networks employing a variety of posterior approximations, such as Gaussian \citep{daxberger2021laplace, yang2023bayesian, blundell2015weight, wang2024blob}, Gaussian mixture \citep{blundell2015weight}, or Dirichlet \citep{hobbhahn2022fast} distributions. In this work, we demonstrate that our PMI score effectively estimates the mutual information of high-dimensional datasets, even when implemented with a simple Bayesian logistic regression.

\subsection{Algorithm and convergence rate}
Combining all the steps, our PMI scoring function for data curation methods can be computed as in~\Cref{alg:curation}. While primarily designed to evaluate a data curation method $f(\cdot)$, practical applications often require assessing a dataset $\widehat D$ that has already been curated. To address this, we extend our method using a heuristic approach: we generate subsampled pairs $(\widehat D_1, T_1), \dots, (\widehat D_k, T_k)$ from the curated and test dataset, then apply the PMI scoring function.

\begin{algorithm}[h!]
\caption{PMI scoring function for data curation method $f(\cdot)$ (or curated dataset $\widehat D$)}
\label{alg:curation}
\begin{algorithmic}[1]
\REQUIRE Datasets $(D_1, T_1), \dots, (D_k, T_k)$ and a data curation method $f(\cdot)$ for evaluation (or a curated dataset $\widehat D$ and a test dataset $T$),
a Bayesian model  parameterized by $\theta\in \Theta$ with tractable posterior approximations $p(\theta|\cdot)$, a value $\eta \in \Theta$.
\ENSURE A score for the curation method $f(\cdot)$
\STATE Apply the curation method $f(\cdot)$ on $D_1, \dots, D_k$ and get the curated datasets $\widehat D_1, \dots, \widehat D_k$. (Or randomly subsample $\widehat D$ and $T$ to get $\widehat D_1, \dots, \widehat D_k$ and  $T_1, \dots, T_k$.)
\STATE For each pair of embedded $(\widehat D_i, T_i)$, train separate Bayesian models on $\widehat D_i$, $T_i$, and their union $\widehat D_i \cup T_i$ to get the posterior probabilities $p(\theta =\eta|\widehat D_i)$, $p(\theta = \eta|T_i)$, and $p(\theta = \eta|\widehat D_i, T_i)$. Compute the pointwise mutual information $U_\eta(\widehat D_i, T_i)$ using \Cref{eqn:MI_score}.
\STATE Return $\frac{1}{k} \sum_{i=1}^k U_\eta(\widehat D_i, T_i)$.
\end{algorithmic}
\end{algorithm}
When the Bayesian model provides accurate posterior probabilities $p(\theta =\eta|\widehat D_i)$, $p(\theta = \eta|T_i)$, and $p(\theta = \eta|\widehat D_i, T_i)$, the algorithm outputs an unbiased estimator of the target metric $I(f(D), T)$, which converges to the true value of $I(f(D), T)$ as $k$ increases.
\begin{corollary} \label{cor:convergence}
	 Assume that the Bayesian model provides accurate posterior probabilities. Then the output of \Cref{alg:curation} provides an unbiased estimator of $I(f(D), T)$. Moreover, assuming that the posteriors are in an exponential family and the datasets have bounded sufficient statistics,\footnote{The assumption of bounded sufficient statistics is similar to that in \citet{belghazi2021minemutualinformationneural}, who assume that the output of the \emph{statistics network} is bounded (see the assumptions in their Theorem 3). When the sufficient statistics are not bounded, we can still obtain a result of the same order by clipping $U_\eta(\cdot)$. See details in \Cref{app:convergence}.}
	 we have $\Pr\left(\left|\frac{1}{k} \sum_{i=1}^k U_\eta(\widehat D_i, T_i) - I(\widehat D,T)\right|  \le \varepsilon \right)\ge 1-\delta$ when $k = O(\log(1/\delta)/\varepsilon^2)$, and we have the expected square error of the estimator decreases as $O(1/k)$.
\end{corollary}

The proof is deferred to \Cref{app:convergence}.
Compared to commonly used MI estimators, our concentration bound is independent of the variable dimensionality, unlike the bound of~\citet{belghazi2021minemutualinformationneural}, which scales as
$
O\left(\frac{d \log(\sqrt{d}/\varepsilon) + d + \log(1/\delta)}{\varepsilon^2}\right),
$
where \(d\) is the variable dimensionality.\footnote{The dependence on $d$ mainly comes from a uniform convergence bound that requires covering a subspace of \(\mathbb{R}^d\) with a finite number of small balls. See their proof of Theorem 6.}
We demonstrate the advantages of our method in~\Cref{sec:exp_accuracy}.

\section{Experiments}

We present a comprehensive evaluation of our proposed PMI estimator, assessing both its accuracy in estimating  dataset mutual information and its effectiveness as a robust scoring function for data curation methods. Our experiments encompass a diverse range of modalities, from standard image classification benchmarks (Colored MNIST, CIFAR-10, ImageNet) to large-scale LLM response preference datasets (Skywork, SHP). Empirically, we find that the PMI score remains highly effective even when implemented using a simple Bayesian logistic regression model (as outlined in~\Cref{app:example}). This allows for efficient posterior approximation via standard L2 regularization or the Laplace approximation method~\citep{daxberger2021laplace}.

We structure our evaluation into three main stages:
\begin{enumerate}
    \item \textbf{Estimation Accuracy (\Cref{sec:exp_accuracy}):} We first verify that our method correctly estimates the mutual information of high-dimensional dataset pairs, significantly outperforming baselines such as MINE and Monte Carlo integration.
    \item \textbf{Robustness to Strategic Curation (\Cref{sec:exp_scoring_evaluation}):} We demonstrate the metric's capability of identifying strategic data curation method (\Cref{def:strategic}).
    \item \textbf{Alignment with Generalization Capability (\Cref{sec:exp_preference}):} Finally, we show that PMI scores align with out-of-distribution (OOD) generalization. In contrast, standard test accuracy favors datasets that resemble the test distribution but fail to generalize to new domains.
\end{enumerate}

\subsection{Accuracy of Mutual Information Estimation} \label{sec:exp_accuracy}
We first evaluate the accuracy of our dataset MI estimator on resampled real-world data.

\paragraph{A benchmark for dataset MI estimation.}
Prior work on MI estimation has primarily focused on pairs of individual data points with dimensionality on the order of $1000$ (e.g., correlated image pairs from CMNIST or CIFAR-10). However, these methods have never been evaluated in settings where the goal is to estimate the mutual information between datasets containing many such data points. To bridge this gap, we introduce a benchmark that extends existing MI evaluation frameworks~\citep{lee2024benchmarksuiteevaluatingneural, gowri2024approximating} to dataset pairs. Our benchmark generates dataset pairs with pre-defined MI values by resampling from standard datasets like CMNIST, CIFAR and ImageNet. We construct datasets with binary labels (e.g., label 0 and 1), each containing both  labels and images. Given a target MI value $\lambda$, we first define a joint distribution $p(r_D, r_T)$ of two correlated numbers $r_D, r_T \in \{0.1, 0.2, \dots, 0.9\}$ such that their mutual information  $I(r_D, r_T) = \lambda$. Here, $r_D$ and $r_T$ represent the proportions of 0-labeled images in datasets $D$ and $T$, respectively. Next, we sample $(r_D, r_T)\sim p(r_D,r_T)$ and generate binary label vectors $L_D\in\{0,1\}^{N_D}$ and $L_T\in\{0,1\}^{N_T}$ whose empirical proportions match $r_D$ and $r_T$. To enforce $H(r_D\mid D)=0$ and $H(r_T\mid T)=0$, we modify a single bit of each label vector so that the dataset deterministically reveals whether its proportion is below or above $0.5$, and then attach an image from the corresponding class to each label to form datasets $D$ and $T$. Under this construction, the ground-truth MI satisfies $I(D,T)=I(r_D,r_T)$. We generate $k=2000$ dataset pairs with sizes ranging from 50 to 100 and target mutual information values spanning $[0,1]$, and evaluate estimation accuracy via Spearman's rank correlation between the estimated and true MI rankings. See~\Cref{app:exp_accuracy} for full details.

\paragraph{Baseline method.} Due to the high dimensionality of datasets, we employ dimensionality reduction prior to evaluation, following standard practice in the neural MI estimation literature. For CMNIST, we use Principal Component Analysis to reduce the feature dimensionality to 100. For CIFAR, we extract embeddings by a ResNet-18 model pretrained on ImageNet, with the final classification layer removed. For ImageNet, we extract embeddings using a ResNet-50 model pretrained on ImageNet, with the final classification layer removed. We then consider the following baselines.
\begin{itemize}
\item \textbf{MINE}: For unstructured high-dimensional data such as images, MINE~\citep{belghazi2021minemutualinformationneural} is one of the most widely-used and effective MI estimators~\citep{gowri2024approximating,lee2024benchmarksuiteevaluatingneural}. According to~\citep{lee2024benchmarksuiteevaluatingneural}, MINE is also the most robust method for image-based data, making it a strong baseline for our setting.
\item \textbf{MINE on trained model parameters $\theta$}: To further reduce dimensionality, we apply MINE not on raw data but on the trained model parameters, which serve as a lower-dimensional proxy.

\item \textbf{LMI}: To further reduce dimensionality, we adopt a recent approach~\citet{gowri2024approximating}, which learns low-dimensional latent representations of the data before estimating mutual information.

\item \textbf{Monte Carlo integration}: We also evaluate a simple Monte Carlo integration approach to compute the pointwise mutual information defined in~\Cref{prop:MI_metric}. Details in~\Cref{app:exp_accuracy}.
\end{itemize}

\paragraph{Our method.} We implement our PMI estimator following~\Cref{alg:curation}.  To obtain the posterior distributions $p(\theta|\cdot)$, we train logistic regression models on embedded $D$ and $T$ with L2 regularization parameterized by $C$. This setup corresponds to Bayesian logistic regression with Gaussian approximation, where the prior is $N(0, C\cdot \mathbf{I})$. See~\Cref{app:example}.

\begin{figure}[htbp]
\centering
    \hspace{-8mm}
    \begin{subfigure}[b]{0.38\textwidth}
        \centering
        \includegraphics[width=\linewidth]{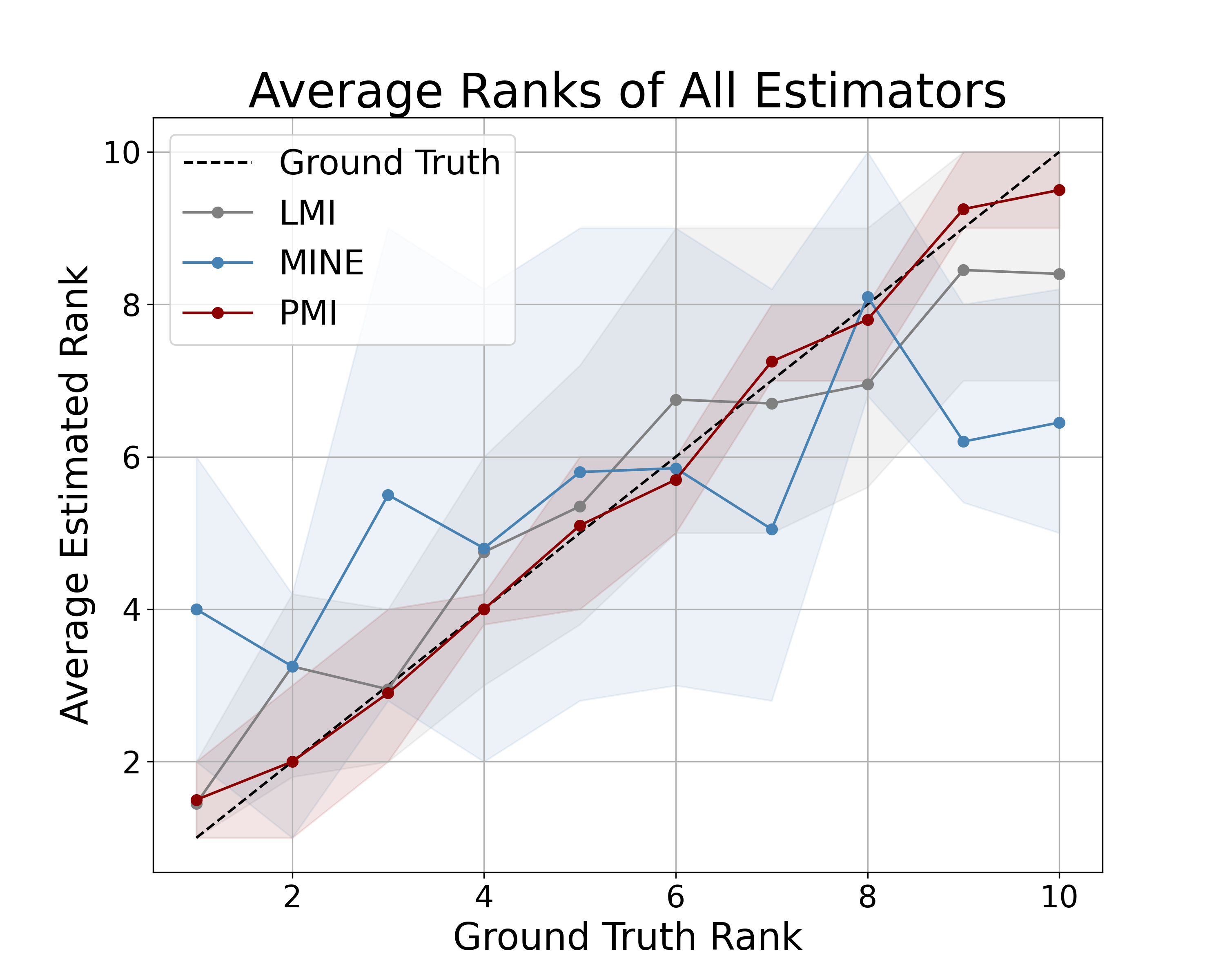}
        \caption{MNIST}
    \end{subfigure}
    \hspace{-8mm}
    \begin{subfigure}[b]{0.38\textwidth}
        \centering
        \includegraphics[width=\linewidth]{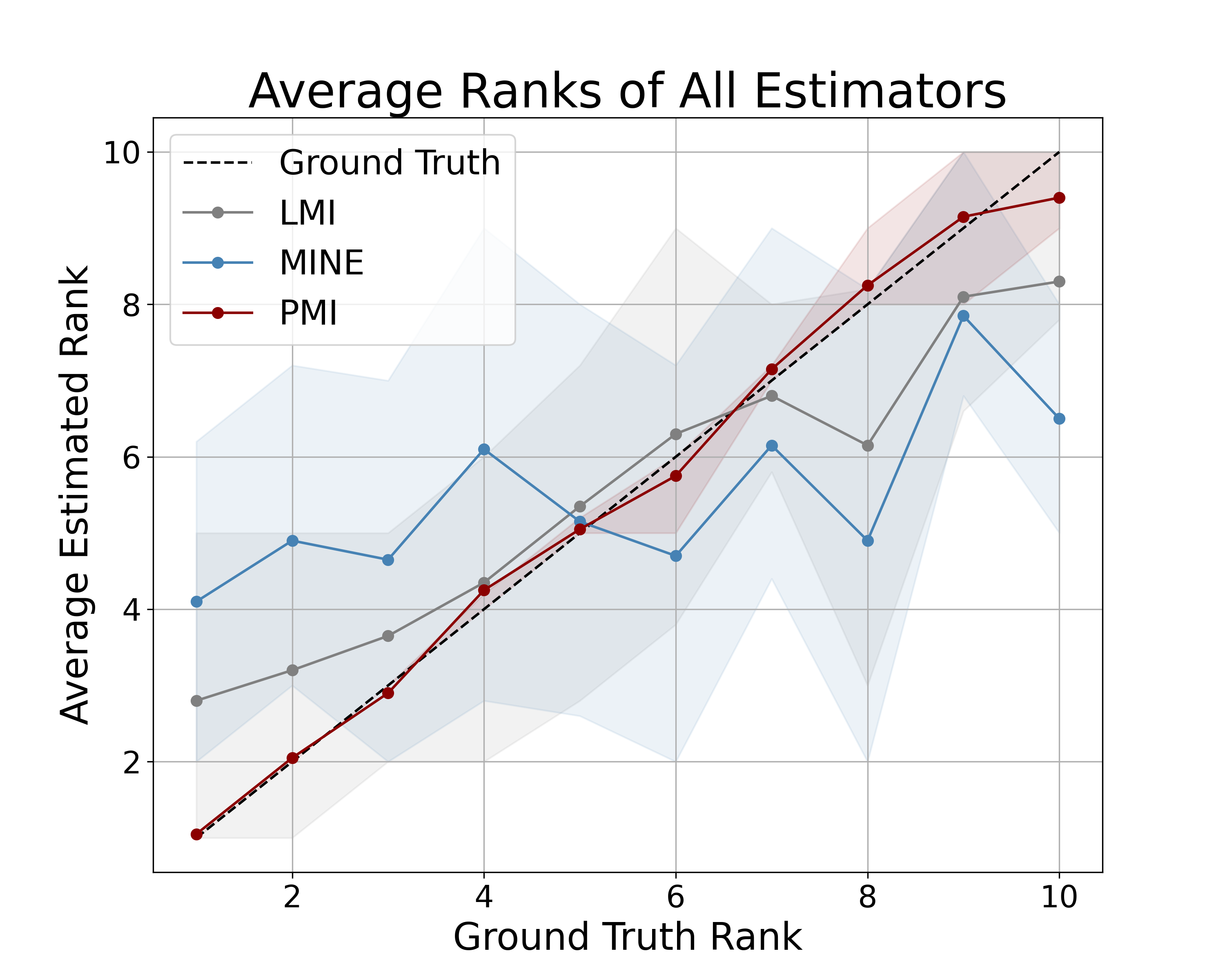}
        \caption{CIFAR}
    \end{subfigure}
    \hspace{-8mm}
    \begin{subfigure}[b]{0.38\textwidth}
        \centering
        \includegraphics[width=\linewidth]{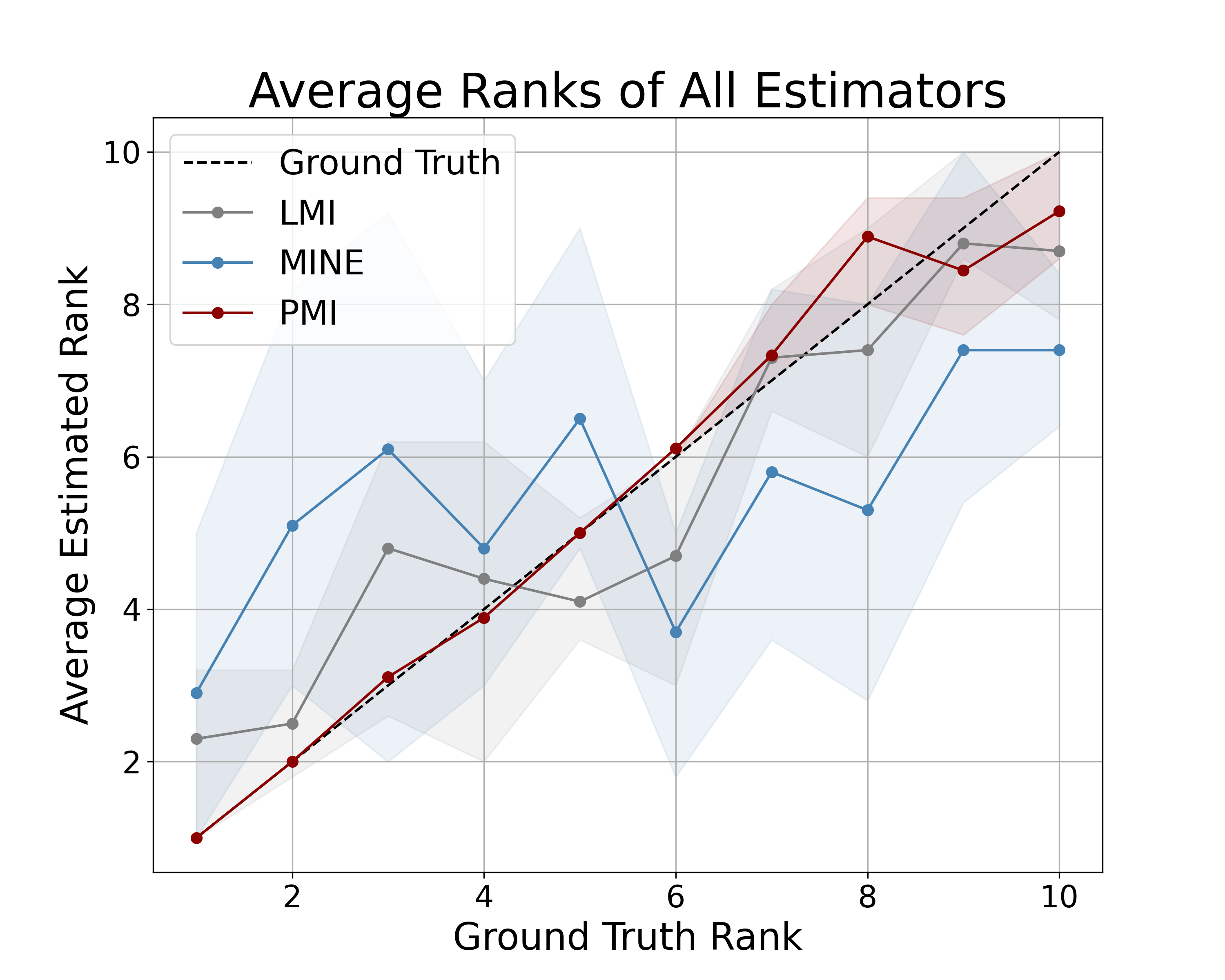}
        \caption{ImageNet}
    \end{subfigure}
    \hspace{-8mm}
    \caption{
    Estimated rankings from different methods on MNIST, CIFAR and ImageNet. PMI produces the most accurate estimates with the smallest variance. The $x$-axis denotes the ground-truth MI ranking indices, and the $y$-axis denotes the estimated rankings generated by each method. The lines represent the mean estimated ranking over 20 independent trials, while the shaded regions indicate the inter-quantile band between the 20th and 80th percentiles across trials.
    The dataset size is $100$.
    }
    \Description{Three ranking plots for MNIST, CIFAR, and ImageNet comparing PMI, MINE, LMI, and Monte Carlo estimates against the ground-truth mutual-information ranking.}
    \label{figure}
\end{figure}

\paragraph{Results.}
As shown in \Cref{tab:tau-and-runtime-vs-method} and \Cref{figure}, our PMI estimator achieves the highest Spearman's \(\rho\) rank correlation, producing the most accurate estimates with the smallest variance.  In addition, in~\Cref{app:exp_accuracy}, we show that PMI outperforms other methods regardless of the choice of regularization strength \(C\), which means its ranking estimates are robust to prior misspecification.  Our method also remains effective across different dataset sizes.

\begin{table}[htbp]
    \centering
    \begin{tabular}{l c c c}
    \hline
    & \textbf{CMNIST} & \textbf{CIFAR} & \textbf{ImageNet} \\
    \cline{2-3}
    \hline
    PMI  & \textbf{0.967}{\scriptsize $\pm$ 0.015} & \textbf{0.976}{\scriptsize $\pm$ 0.000} & \textbf{0.972}{\scriptsize $\pm$ 0.021} \\
    \hline
    MINE & 0.354{\scriptsize $\pm$ 0.022} & 0.281{\scriptsize $\pm$ 0.040} & {0.350}{\scriptsize $\pm$ 0.023} \\
    MINE on  $\theta$ & 0.367{\scriptsize $\pm$ 0.033} & 0.312{\scriptsize $\pm$ 0.043} & {0.405}{\scriptsize $\pm$ 0.036} \\
    LMI & 0.765{\scriptsize $\pm$ 0.174} & 0.634{\scriptsize $\pm$ 0.033} & 0.752{\scriptsize $\pm$ 0.143}\\
        Monte Carlo & 0.611 {\scriptsize $\pm$ 0.029}  & Failed & Failed \\
    \hline
    \end{tabular}
    \caption{
        Spearman's rank correlation (\(\rho\)) between estimated and ground-truth mutual information rankings for different estimation methods on Colored MNIST~\citep{arjovsky2020invariantriskminimization}, CIFAR~\citep{krizhevsky2009learning} and ImageNet~\citep{deng2009imagenet}. Higher values indicate better alignment with the true MI ranking. PMI achieves the strongest correlation. We run 20 independent trials to compute the standard deviation, which quantifies the variability of estimation outcomes across repeated measurements. The Monte Carlo method fails on CIFAR, consistently outputting zero due to numerical instability caused by near-zero likelihoods.
        The dataset size is $100$.
    }
    \label{tab:tau-and-runtime-vs-method}
\end{table}

\subsection{Evaluating data curation methods} \label{sec:exp_scoring_evaluation}
We next test our PMI scoring function in evaluating data curation methods (outlined as PMI scoring function for data curation methods in \Cref{alg:curation}). We show that the PMI scoring function is effective in distinguishing between strategic and non-strategic curation methods, whereas evaluating curation methods using test scores could promote strategic methods that do not add new information but merely make the data more similar to the test data.

\begin{figure}[t]
    \centering
    \begin{subfigure}[b]{0.32\textwidth}
        \centering
    \includegraphics[width=0.48\linewidth]{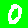}\hfill
        \includegraphics[width=0.48\linewidth]{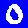}
        \caption{Colored MNIST}
        \label{fig:cmnist}
    \end{subfigure}
    \hfill
    \begin{subfigure}[b]{0.32\textwidth}
        \centering
        \includegraphics[width=0.48\linewidth]{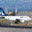}\hfill
        \includegraphics[width=0.48\linewidth]{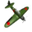}
        \caption{Corrupted CIFAR}
        \label{fig:cifar}
    \end{subfigure}
    \hfill
    \begin{subfigure}[b]{0.32\textwidth}
        \centering
        \includegraphics[width=0.48\linewidth]{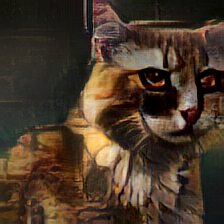}\hfill
        \includegraphics[width=0.48\linewidth]{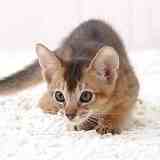}
        \caption{Stylized ImageNet}
        \label{fig:imagenet}
    \end{subfigure}
    \caption{Visual samples from the three benchmark datasets with different non-essential feature $z_N$, including color, picture corruption, and picture style.
    }
    \Description{Example image pairs from Colored MNIST, Corrupted CIFAR, and Stylized ImageNet showing changes in non-essential color, corruption, and style attributes.}
    \label{fig:all_datasets}
\end{figure}

\paragraph{Problem Setup and Datasets}
We formally define the data generation process for our evaluation. We focus on image datasets and each image input $\mathbf{x}$ contains essential feature $\mathbf{z}_E$ that is indicative of label $y$ and a non-essential feature $z_N$ that is irrelevant to the label.
We evaluate our method with resampled data from three datasets, with dataset sizes ranging from 80 to 480. \Cref{fig:all_datasets} provides visual illustrations of these benchmarks and the corresponding non-essential features $z_N$ (color, corruption, and style).
\begin{itemize}
    \item \textbf{Colored MNIST}~\citep{arjovsky2020invariantriskminimization}: The task is binary classification of digits. Here, $\mathbf{z}_E$ represents the digit shape (``0'' vs. ``1''), while $z_N$ corresponds to the background color, which serves as a non-essential feature.
    \item \textbf{Corrupted CIFAR}~\citep{hendrycks2019benchmarkingneuralnetworkrobustness}: We select ``Airplane'' and ``Automobile'' as the target labels. In this setting, $\mathbf{z}_E$ includes the object semantics, and $z_N$ represents the corruption type applied to the image (``Brightness'' vs. ``Contrast'').
    \item \textbf{Stylized ImageNet}~\citep{geirhos2022imagenettrainedcnnsbiasedtexture}: The binary targets are ``Cat'' and ``Dog''. $\mathbf{z}_E$ includes the animal's shape and texture, while $z_N$ refers to the distinct artistic style used to render the image.
\end{itemize}

\paragraph{Data curation methods} There are numerous data curation methods available for evaluation. We select two data curation methods that can be clearly categorized as strategic or non-strategic, allowing us to verify whether they are correctly classified.
\begin{itemize}
	\item \textbf{Data filtering:} We consider the removal of mislabeled data, a non-strategic curation method that is expected to yield a higher score than leaving the data unmodified.
	We randomly sample $T_1, \dots, T_k$ from the test set, and sample datasets from the training set and flip the labels of some data points to generate $D_1, \dots, D_k$. We compare the scores before and after the removal of these mislabeled data points. We clarify this as a \textit{non-strategic} curation method because the filtering criterion is grounded in the intrinsic quality of the data (i.e., correcting label noise based on the ground truth) rather than strategically aligning with the test distribution statistics.

	\item \textbf{Strategic data removal by non-essential features:}
	We next consider a strategic curation method that aligns the training distribution with the test distribution strictly through data removal based solely on non-essential features. We consider the setting where the core semantic distribution is consistent across domains, i.e., $p_D(\mathbf{z}_E, y) \approx p_T(\mathbf{z}_E, y)$, but the conditional distribution of the non-essential feature shifts: $p_D(z_N \mid \mathbf{z}_E, y) \neq p_T(z_N \mid \mathbf{z}_E, y)$. To address this, we construct a curated dataset $\hat{D} \subset D$ by subsampling the training data. Specifically, we remove examples from $D$ to force the empirical conditional distribution $p_{\hat{D}}(z_N \mid \mathbf{z}_E, y)$ to match that of the target test set $T$. We classify this as a \textit{strategic curation method} because the removal decision is conditioned solely on the distributional statistics of $(D, T)$ and is independent of the true model parameter $\theta$. Since this method reduces the effective sample size and disregards individual data quality, we hypothesize that it will yield lower performance compared to the full dataset baseline, $f(D) \equiv D$.
\end{itemize}
For both cases, we generate the smallest datasets that achieve reasonable accuracy $\sim 80\% - 90\%$ to avoid overlap. Crucially, we ensure that the final dataset size is identical for both methods to allow for a fair comparison.

\paragraph{Scoring functions.}
We apply the PMI scoring function for data curation methods outlined in \Cref{alg:curation}. We compare our PMI dataset score  with the test-score-based evaluation defined as:
\(
S_{TS}(f(\cdot)) = \frac{1}{k} \sum_{i=1}^k \text{Acc}(\hat\theta(f(D_i)), T_i),
\)
where \( \text{Acc}(\hat\theta(D), T) \) denotes the accuracy of a model trained on \( D \) and evaluated on \( T \). To compute the PMI score, we train a standard logistic regression for Colored MNIST, while for Corrupted CIFAR and Stylized ImageNet, we train logistic regression on embeddings extracted from ResNet-18 and ResNet-50 backbones, respectively. As in~\Cref{sec:exp_accuracy}, all logistic regression models are trained with L2 regularization parameter \( C \), corresponding to a Gaussian prior \( N(0, C \cdot \mathbf{I}) \). These models are also used to compute test accuracy.

\begin{table}
\centering
\begin{tabular}{cccc}
\toprule
\textbf{Dataset} & \textbf{Operation} & \textbf{$\Delta$ PMI Score } & \textbf{$\Delta$ Test Accuracy (\%)} \\
\midrule
\multirow{2}{*}{Colored MNIST}
& \textcolor{accessblue}{Filtering}   & \textcolor{accessblue}{13.59} \scriptsize{$\pm$ 1.04} & \textcolor{accessblue}{0.31} \scriptsize{$\pm$ 0.02} \\
& \textcolor{accessorange}{Removal}     & \textcolor{accessorange}{-13.96} \scriptsize{$\pm$ 2.05} & \textcolor{accessblue}{0.54} \scriptsize{$\pm$ 0.04} \\
\midrule
\multirow{2}{*}{Corrupted CIFAR}
& \textcolor{accessblue}{Filtering}   & \textcolor{accessblue}{1.53} \scriptsize{$\pm$ 0.15} & \textcolor{accessblue}{7.29} \scriptsize{$\pm$ 0.10} \\
& \textcolor{accessorange}{Removal}     & \textcolor{accessorange}{-4.68} \scriptsize{$\pm$ 0.12} & \textcolor{accessblue}{0.82} \scriptsize{$\pm$ 0.13} \\
\midrule
\multirow{2}{*}{Stylized ImageNet}
& \textcolor{accessblue}{Filtering}   & \textcolor{accessblue}{20.95} \scriptsize{$\pm$ 0.15} & \textcolor{accessblue}{2.74} \scriptsize{$\pm$ 0.09} \\
& \textcolor{accessorange}{Removal}     & \textcolor{accessorange}{-2.75} \scriptsize{$\pm$ 0.13} & \textcolor{accessblue}{1.00} \scriptsize{$\pm$ 0.07} \\
\midrule
\multicolumn{2}{c}{\textbf{Strategy-proof}} & Yes & No \\
\bottomrule
\end{tabular}
\caption{Change in PMI score and test accuracy \( S_{TS} \) after applying two data curation methods to Colored MNIST, Corrupted CIFAR and Stylized ImageNet datasets. The PMI score effectively distinguishes between \textcolor{accessorange}{strategic} and \textcolor{accessblue}{non-strategic} methods: \textcolor{accessblue}{Filtering} increases the PMI score, while \textcolor{accessorange}{Removal} based on non-essential features decreases it. We ensure that both methods discard the \textbf{exact same number} of data points to maintain a fair comparison. The dataset sizes range from $80$ to $480$.
 In contrast, test accuracy fails to detect \textcolor{accessorange}{strategic} methods, always assigning higher scores. The experiment is repeated 1,000 times to compute mean changes in PMI scores and accuracy and repeated 10 times to obtain final means and variances. Details of the experimental setup are provided in \Cref{app:exp_curation}.}
\label{tab:combined-results}
\end{table}

\begin{figure}[ht]
    \centering
    \includegraphics[width=0.7\textwidth]{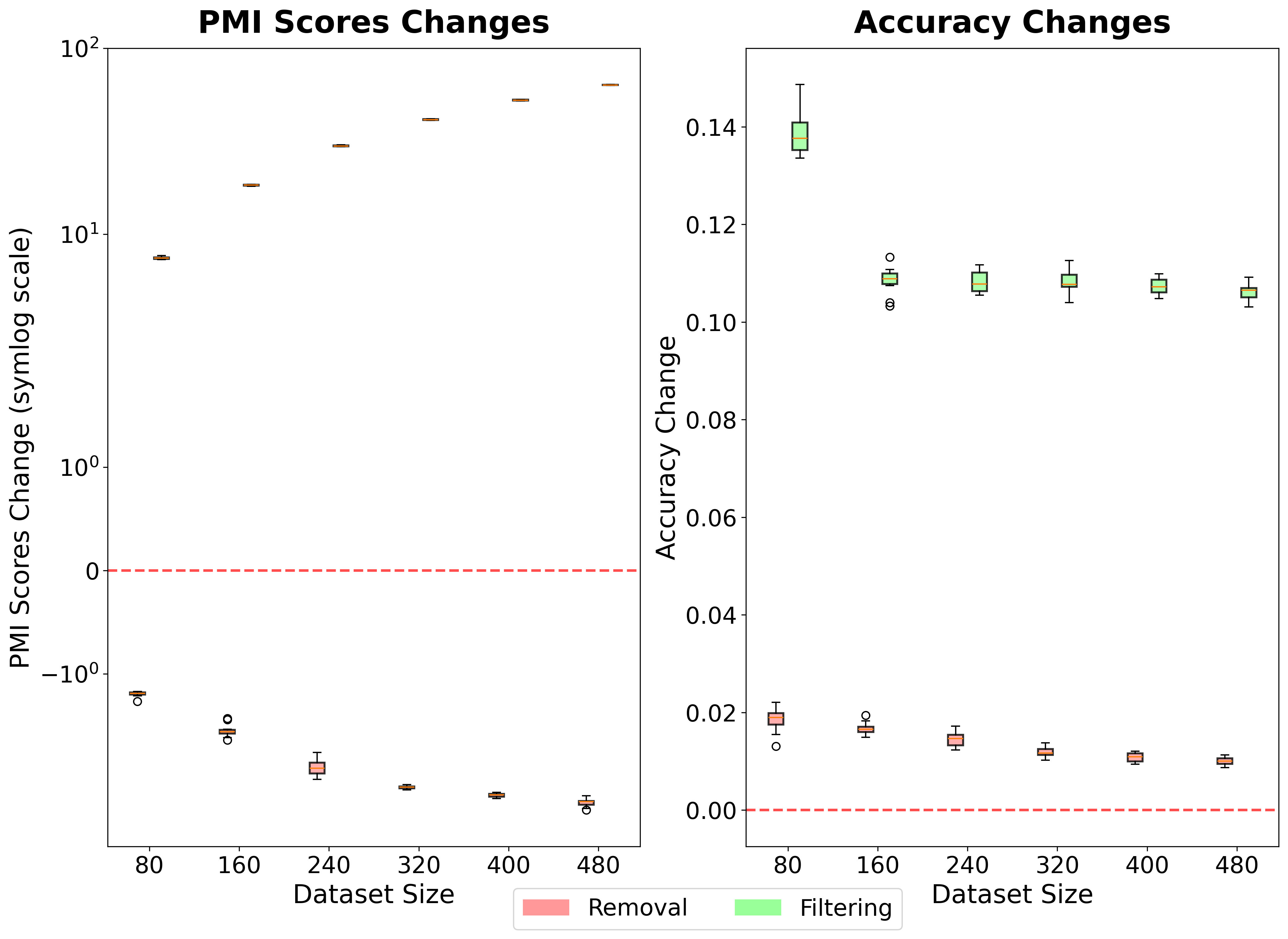}
    \caption{Visual representation of the changes in PMI score and test accuracy on Stylized ImageNet as a function of dataset size (80 to 480 samples). The dashed horizontal line represents the baseline performance of the original, unmodified dataset. Both Filtering and Removal are shown in both panels. In the left panel, the \textbf{PMI Score} (symlog scale) sharply separates the two methods: \textcolor{accessblue}{Filtering} improves the score, while \textcolor{accessorange}{Removal} based on non-essential features significantly degrades it. The right panel shows \textbf{Test Accuracy}, where both methods yield similar improvements over the baseline, failing to reflect the strategic nature of the removal method. Boxplots represent the variance across 10 repetitions.}
    \Description{Boxplots comparing changes in PMI score and test accuracy for Filtering and Removal on Stylized ImageNet across dataset sizes.}
    \label{fig:stylized_results_plot}
\end{figure}

\begin{table}[ht]
    \centering
    \begin{tabular}{ccccc}
        \toprule
        \multirow{2}{*}{\textbf{C}} & \multicolumn{2}{c}{\textbf{PMI Score Change}} & \multicolumn{2}{c}{\textbf{Acc. Change (\%)}} \\
        \cmidrule(lr){2-3} \cmidrule(lr){4-5}
         & \textbf{\textcolor{accessblue}{Filtering}} & \textbf{\textcolor{accessorange}{Removal}} & \textbf{\textcolor{accessblue}{Filtering}} & \textbf{\textcolor{accessorange}{Removal}} \\
        \midrule
        $10000$  & \textcolor{accessblue}{$2.12\pm0.19$} & \textcolor{accessorange}{$-6.17\pm0.06$} & \textcolor{accessblue}{$7.38\pm0.11$} & \textcolor{accessblue}{$1.83\pm0.12$} \\
        $30000$  & \textcolor{accessblue}{$1.99\pm0.21$} & \textcolor{accessorange}{$-7.31\pm0.14$} & \textcolor{accessblue}{$7.36\pm0.15$} & \textcolor{accessblue}{$1.85\pm0.12$} \\
        $50000$  & \textcolor{accessblue}{$1.83\pm0.13$} & \textcolor{accessorange}{$-7.84\pm0.11$} & \textcolor{accessblue}{$7.26\pm0.14$} & \textcolor{accessblue}{$1.86\pm0.14$} \\
        $100000$ & \textcolor{accessblue}{$1.58\pm0.17$} & \textcolor{accessorange}{$-8.36\pm0.13$} & \textcolor{accessblue}{$7.31\pm0.09$} & \textcolor{accessblue}{$1.92\pm0.09$} \\
        \bottomrule
    \end{tabular}
    \caption{\textbf{Changes} in PMI score function and test accuracy after applying data curation methods in Corrupted CIFAR dataset.  Here, $C$ represents the L2 regularization parameter of the trained logistic regression, corresponding to a Gaussian prior $N(0, C \cdot \mathbf{I})$. The results demonstrate that our PMI scores are robust to prior misspecification. The PMI score consistently distinguishes between \textcolor{accessorange}{strategic (Removal)} and \textcolor{accessblue}{non-strategic (Filtering)} methods. In contrast, test accuracy consistently fails to detect \textcolor{accessorange}{strategic data removal}.}
    \label{tab:cifar-filter-remove}
\end{table}

\paragraph{Results.}
\Cref{tab:combined-results}, \Cref{tab:cifar-filter-remove} and \Cref{fig:stylized_results_plot} report changes in PMI score and test accuracy \( S_{TS} \) after applying two data curation methods. The PMI score effectively distinguishes between strategic and non-strategic data curation methods: data filtering increases the score, while removal based on non-essential features decreases it. In contrast, test accuracy fails to detect strategic data curation methods, consistently assigning higher scores. This highlights the risk of relying solely on test accuracy, which may inadvertently promote strategic methods that do not introduce new information but merely make the training data more similar to the test data. Our results are robust to prior misspecification (i.e., different choices of regularization parameter $C$) and are consistent across different data distributions, as detailed in \Cref{app:exp_curation}.

\subsection{Aligning with out-of-distribution generalization}
\label{sec:exp_preference}

In this section, we evaluate our PMI scoring function for curated datasets outlined in \Cref{alg:curation}. We investigate the relationship between our proposed PMI score and model generalization capabilities across datasets with varying degrees of diversity. Recent research increasingly identifies data diversity as a key factor in improving model robustness and out-of-distribution (OOD) generalization~\citep{jung2025prismatic, zhou2023limaalignment, bukharin-etal-2024-data, pang2025improvingdataefficiencycurating}. Specifically, \citet{jung2025prismatic} demonstrate that diversity metrics based on loss gradients strongly predict reasoning generalization.
Building on this intuition, we hypothesize that a robust dataset evaluation metric should favor diverse datasets that generalize well to external benchmarks, rather than homogeneous datasets that simply overfit the distributional patterns of a specific test set. \textbf{We empirically demonstrate that while standard test accuracy is misled by redundant, low-diversity datasets, our PMI score correctly aligns with true out-of-distribution (OOD) generalization, effectively distinguishing high-quality diverse data from those that simply overfit the test distribution.}

\paragraph{LLM Datasets.} Modern Large Language Models (LLMs) rely heavily on Reinforcement Learning from Human Feedback (RLHF) to align model outputs with human intent. A critical component of this framework is \textit{instruction-following}, the capability of an agent to accurately understand and execute specific user queries  rather than merely completing text patterns. This alignment process typically requires large-scale preference datasets, where a reward model learns to distinguish between desirable (``chosen'') and undesirable (``rejected'') responses to the same prompt.

In this work, we focus on the text-based preference datasets used for this specific LLM RLHF stage. For our primary data source, we utilize the \textbf{Skywork-Reward-Preference-80K} dataset~\citep{liu2024skyworkrewardbagtricksreward}. Specifically, we utilize the \textit{Magpie-derived} subsets, which provide a large-scale repository of high-quality synthetic instruction-following data. These subsets comprise approximately 55,000 samples covering technical domains like mathematics and coding. \textbf{Crucially, the preference data here is defined as a comparative ranking between two candidate outputs given a fixed context.} Each sample consists of a single user prompt (the instruction) and two distinct model-generated responses. The preference label indicates which response better satisfies the prompt---identifying the ``chosen'' response over the ``rejected'' counterpart, where the chosen response is typically optimized for structural correctness and detailed derivation.

From this pool, we randomly hold out 5,000 samples as the \textit{Test Set} (Skywork Test) to compute in-distribution test accuracy and PMI. The remaining $\sim$50,000 samples constitute the candidate pool for our training data selection experiments.

To evaluate true generalization capabilities, we employ the \textbf{Stanford Human Preferences (SHP)} dataset~\citep{ethayarajh2025understandingdatasetdifficultymathcalvusable} as an \textit{external OOD benchmark}. In contrast to Skywork, SHP consists of naturally occurring human judgments sourced from 18 diverse subreddits (e.g., \textit{askculinary}, \textit{legaladvice}, \textit{askacademia}). The preference labels in SHP are determined by community consensus rather than heuristic rules or model-based scoring. This introduces a significant distributional shift: models must transfer knowledge from the highly structured, formal domain of synthetic data to the informal, context-dependent nature of real-world human preferences.

\paragraph{Gradient-based Diversity Sampling.}

To construct training sets with distinct diversity profiles, we adopt the gradient-based representation method  \citet{jung2025prismatic}, positing that loss gradients capture the underlying learning process more effectively than semantic embeddings. We utilize the \texttt{Qwen3-1.7B} model~\citep{qwen3} to derive these representations. To manage the high dimensionality of the parameter space ($|\theta| \approx 1.7$B), we apply a \textbf{Rademacher random projection}~\citep{press2007numerical, rakhshan2020tensorized} to compress the gradients into a target dimension of $d=1024$. The input query-response pairs are formatted with a specific task instruction to guide the gradient extraction towards quality and helpfulness features.

Denote a preference pair as $(q, r_w, r_l)$, where $q$ represents the user query, $r_w$ is the chosen response, and $r_l$ is the rejected response. We compute the gradients of this loss with respect to the model parameters separately for the chosen response $r_w$ and the rejected response $r_l$. Finally, these two projected gradient vectors are concatenated to form a single, unified embedding vector $\mathbf{g} = [\nabla_{\theta}^{\text{proj}} \mathcal{L}(q,r_w) ; \nabla_{\theta}^{\text{proj}} \mathcal{L}(q,r_l)]$ for each data pair. See Appendix~\ref{app:gradient_details} for the exact task instruction template and mathematical formulation.

Let $G = \{g_1, \dots, g_N\}$ denote the gradient-based embeddings of the candidate pool. Based on these representations, we sample three types of datasets (10 independent subsets per type, each containing 5,000 samples):

\begin{itemize}
    \item \textbf{Distribution Smoothing:} To maximize the semantic coverage of the selected subset, we employ a clustering-based stratified sampling method. We first apply K-Means clustering to partition the embeddings $G$ into $k$ distinct clusters ($C_1, \dots, C_k$). We then perform uniform sampling across these clusters, selecting an equal number of samples from each partition. This approach approximates uniform geometric coverage within the gradient space. By effectively flattening the data density, it ensures the retention of rare reasoning patterns and outliers that are typically marginalized in the natural distribution.

    \item \textbf{Peak Selection:} To simulate a dataset characterized by high redundancy and limited semantic range, we utilize a greedy similarity-based algorithm. The selection process is initialized with a randomly sampled seed $s \in G$. In each iteration, we calculate the cosine similarity between the remaining candidates and the current selected set $S$. We exclusively select candidates that exhibit high similarity (threshold $\tau > 0.8$) to any sample in $S$. This greedy accumulation forms a sharp density peak of repetitive data, effectively simulating a scenario of high confidence but low novelty.

    \item \textbf{Uniform Random:} As a representation of the inherent properties of the source distribution, we perform uniform random sampling from the candidate pool $G$. Unlike \textit{Distribution Smoothing}, which flattens the density, or \textit{Peak Selection}, which artificially concentrates it, this approach preserves the original density and biases of the \textit{Magpie-derived} source pool, serving as a reference point for the natural diversity level.
\end{itemize}

By construction, these techniques induce a strict diversity hierarchy: \textbf{Distribution Smoothing} exhibits the maximal semantic coverage, followed by the natural baseline of \textbf{Uniform Random}, while \textbf{Peak Selection} yields the lowest diversity due to its artificial redundancy.

\paragraph{Experimental Setup.}
To comprehensively evaluate the curated datasets, we adopt two distinct evaluation paradigms: an information-theoretic metric via a linear proxy model, and an empirical metric via Supervised Fine-Tuning (SFT).
\begin{itemize}
    \item \textbf{PMI Score:}  We apply the PMI scoring function outlined in \Cref{alg:curation}, where the test set is the \textit{Observed Skywork Test Set}. To formulate PMI estimation as a tractable classification problem, we construct balanced binary labels from the raw preference pairs. For each preference pair $(q, r_w, r_l)$, we assign a synthetic label $y \sim \text{Bernoulli}(0.5)$. The input feature vector $\mathbf{x}$ is then defined by the concatenated gradients conditioned on $y$:
        \begin{equation*}
            \mathbf{x} =
            \begin{cases}
            [\nabla_{\theta}^{\text{proj}} \mathcal{L}(q,r_w) ; \nabla_{\theta}^{\text{proj}} \mathcal{L}(q,r_l)] & \text{if } y = 0, \\
            [\nabla_{\theta}^{\text{proj}} \mathcal{L}(q,r_l) ; \nabla_{\theta}^{\text{proj}} \mathcal{L}(q,r_w)] & \text{if } y = 1.
            \end{cases}
        \end{equation*}
     Then we employ Logistic Regression with an L2 regularization strength of $C=0.001$ to estimate PMI. For each dataset, we subsample 1,000 data points from both the training and test sets, compute the PMI, and repeat this process 100 times to report the mean.
    \item \textbf{Empirical Accuracies:} To assess practical downstream impact, we independently fine-tune the \texttt{Qwen3-1.7B} model as a Reward Model on each curated dataset:
    \begin{itemize}
        \item \textbf{Test Accuracy (In-Distribution):} Accuracy on the \textit{Observed Skywork Test Set}, measuring the model's alignment with the synthetic source data's specific stylistic and formatting patterns.
        \item \textbf{Generalization Accuracy (OOD):} Accuracy on the \textit{External OOD Set (SHP)}. Sourced from natural Reddit queries, this metric evaluates true reasoning robustness rather than structural overfitting.
    \end{itemize}
\end{itemize}

\begin{table}[h]
\centering
\resizebox{\linewidth}{!}{
\begin{tabular}{cccc}
\toprule
\textbf{Sampling Strategy} & \textbf{PMI score} & \textbf{SFT acc on SHP (\%)} & \textbf{SFT acc on Skywork Test (\%)} \\
\midrule
Distribution Smoothing & \textcolor{accessblue}{40.97} \scriptsize{($\pm$ 0.25)} & \textcolor{accessblue}{57.60} & \textcolor{accessorange}{92.16} \\
Peak Selection & \textcolor{accessorange}{25.80} \scriptsize{($\pm$ 0.27)} & \textcolor{accessorange}{57.12} & \textcolor{accessblue}{92.61} \\
Uniform Random & 40.49 \scriptsize{($\pm$ 0.26)} & 57.48 & 92.24 \\
\bottomrule
\end{tabular}
}
\caption{Quantitative comparison of PMI, Skywork Test Accuracy, and SHP Generalization Accuracy.
Highlighted entries indicate the best performance (or highest score) and the worst performance; the values and strategy names identify the ordering without relying on color. The values in parentheses denote the mean standard deviation of PMI scores, computed over 100 iterations. We observe a \textit{performance inversion}: The \textbf{Peak Selection} strategy achieves the highest Skywork Test Accuracy but the lowest Generalization Accuracy (SHP). Crucially, our \textbf{PMI score} aligns with the generalization ranking, effectively penalizing the low-diversity Peak Selection strategy despite its increased test accuracy.
}
\label{tab::diversity-results}
\end{table}

\begin{figure}[h]
    \vspace{-1.5em}
    \centering
    \includegraphics[width=0.8\linewidth]{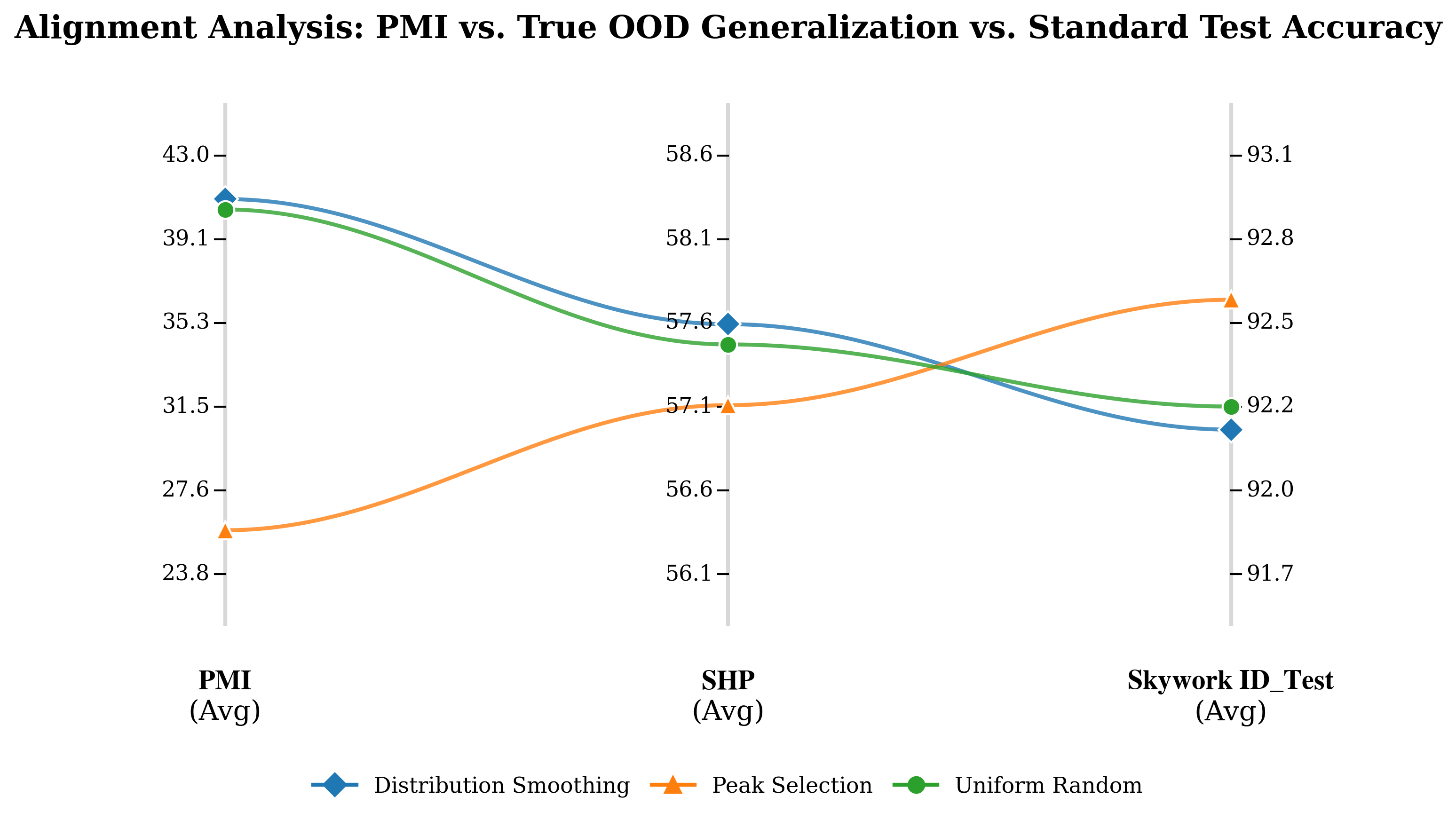}
    \caption{
    \textbf{Alignment Analysis between PMI, Generalization Accuracy (SHP), and Skywork Test Accuracy.} The plot ranks three sampling strategies across the three metrics. On the left (PMI $\to$ SHP), no line crossings indicate strict monotonic alignment, preserving the order $\text{Distribution Smoothing} \succ \text{Uniform Random} \succ \text{Peak Selection}$. On the right (SHP $\to$ Skywork), a sharp crossover marks a performance inversion: the \textcolor{accessorange}{Peak Selection} strategy achieves the highest test accuracy despite the lowest generalization, while the \textcolor{accessblue}{Distribution Smoothing} strategy drops. This shows how standard test accuracy can be inversely correlated with generalization when data diversity is compromised.
    }
     \Description{A parallel-coordinate plot comparing the rankings of Distribution Smoothing, Uniform Random, and Peak Selection under PMI score, SHP generalization accuracy, and Skywork test accuracy.}
     \label{fig::alignment_analysis}
\end{figure}

\paragraph{Results and Analysis.}
As shown in Table~\ref{tab::diversity-results} and Figure~\ref{fig::alignment_analysis},
standard \textit{Skywork Test Accuracy} fails to identify data curation methods that harm model generalization: it incorrectly identifies the \textbf{Peak Selection} strategy as superior (\textcolor{accessblue}{92.61\%}) despite its performance collapse on the out-of-distribution SHP benchmark (\textcolor{accessorange}{57.12\%}).
This evidence suggests that test accuracy may favor datasets that overfit to common patterns in test data, rather than encouraging improvements that truly enhance diversity and generalization.
In contrast, \textit{PMI score}
identifies \textbf{Distribution Smoothing} as the most robust strategy (Score: \textcolor{accessblue}{40.97}), aligning perfectly with the superior generalization observed on SHP (\textcolor{accessblue}{57.6\%}). This strict monotonic alignment ($\text{PMI} \propto \text{Generalization}$) validates PMI as a good proxy for model generalization, capable of distinguishing genuine data improvement from strategic distributional alignment where standard metrics fail.

\section{Discussion and Future Work} \label{sec:discussion}
We propose an information-theoretic framework for evaluating data curation methods that measures data informativeness  by the mutual information of the curated data and the test data. We discuss several potential directions for future work.
Firstly, a key open problem is to develop a principled method for selecting dataset pairs
that most effectively estimate mutual information. We have observed that the PMI scoring function can fail when the datasets $D_i, T_i$ are too small to train effective models, as well as when they are too large, resulting in significant overlap between datasets that violates the independence assumption.
Secondly, the selection of the prior is also crucial. While we observe that the PMI scoring function is robust to prior misspecifications in terms of ranking mutual information, the absolute accuracy of its MI estimates is highly sensitive to the choice of prior.
Thirdly, our experiments focus on the simple logistic regression for Bayesian modeling. It remains an open question whether mutual information estimation could be improved by  more advanced Bayesian neural networks.

\newpage
\bibliographystyle{plainnat}
\bibliography{reference}

\newpage
\appendix

\section{Blackwell ordering} \label{app:blackwell}

We begin by providing background on the Blackwell order of information structures. We first introduce the formal definitions of decision-making problems and information structures.

\begin{definition}
A decision-making problem under uncertainty is defined by the following components:
\begin{itemize}
    \item \textbf{State Space} ($\Omega$): A set of possible states of the world, denoted $\omega \in \Omega$.
    \item \textbf{Action Space} ($A$): A set of possible actions or decisions, denoted $a \in A$.
    \item \textbf{Utility Function} ($u$): A function $u: A \times \Omega \to \mathbb{R}$ that quantifies the payoff of taking action $a$ in state $\omega$.
    \item \textbf{Prior Belief} ($P$): A probability distribution over $\Omega$, representing the decision-maker's initial beliefs. And the corresponding random variable for the state is denoted by $W$.
\end{itemize}
\end{definition}
An information structure reveals some signal about the state of the world $\omega$.
\begin{definition}
    An information structure $S$ consists of a pair $(\mathcal{Y}, \pi)$, where:
        \begin{itemize}
            \item $\mathcal{Y}$ is a set of possible signals or observations.
            \item $\pi: \Omega \to \Delta(\mathcal{Y})$ is a Markov kernel specifying the conditional probability $\pi(y|\omega)$ of observing signal $y$ given state $\omega$.  The corresponding random variable representing the signal is denoted by $Y$.
        \end{itemize}
\end{definition}
The decision-maker observes a signal $y$ from the information structure and updates their beliefs about the state $\omega$ using Bayes' rule. Based on the updated beliefs, they choose an action $a$ to maximize their expected utility.

The Blackwell order provides a way to compare two information structures in terms of their informativeness, which is defined as follows.
\begin{definition}[\citet{Experiments:112749}]
Let $S_1 = (\mathcal{Y}_1, \pi_1)$ and $S_2 = (\mathcal{Y}_2, \pi_2)$ be two information structures over a common state space $\Omega$, with the corresponding signals represented by random variables $Y_1$ and $Y_2$. We say that $S_1$ is \emph{more informative} than $S_2$ in the Blackwell order, if there exists a Markov kernel $\kappa: \mathcal{Y}_1 \to \Delta(\mathcal{Y}_2)$ such that:
\[
\pi_2(y_2|\omega) = \sum_{y_1 \in Y_1} \kappa(y_2|y_1) \pi_1(y_1|\omega) \quad \forall y_2 \in \mathcal{Y}_2, \omega \in \Omega,
\]
or equivalently
$
W \to Y_1 \to Y_2
$
forms a Markov chain, where $W$ is the random variable representing the state.
\end{definition}
In particular, if an information structure $S_1$ is more informative than $S_2$ in the Blackwell order, then, by Blackwell's theorem on decision-making superiority, the decision-maker can achieve at least as high an expected utility using $S_1$ as they can using $S_2$ for any decision-making problem.
\begin{theorem}[Blackwell's theorem on decision-making superiority~\citep{Experiments:112749}]
Let $S_1 = (\mathcal{Y}_1, \pi_1)$ and $S_2 = (\mathcal{Y}_2, \pi_2)$ be two information structures over a common state space $\Omega$ with the corresponding signals represented by random variables $Y_1$ and $Y_2$. The following statements are equivalent:
\begin{enumerate}
    \item \textbf{Blackwell Informativeness}: $S_1$ is more informative than $S_2$, or equivalently, $W\to Y_1 \to Y_2$  forms a Markov chain.
    \item \textbf{Decision-Making Superiority}: For any decision-making problem $(\Omega, A, u, P)$, the maximum expected utility achievable using $S_1$ is at least as high as that achievable using $S_2$. Formally:
        \[
        \max_{a_1: \mathcal{Y}_1 \to A} \mathbb{E}[u(a_1(y_1), \omega)] \geq \max_{a_2: \mathcal{Y}_2 \to A} \mathbb{E}[u(a_2(y_2), \omega)],
        \]
        where the expectations are taken over $\omega \sim P$, $y_1 \sim \pi_1(\cdot|\omega)$, and $y_2 \sim \pi_2(\cdot|\omega)$.
\end{enumerate}
\end{theorem}

We can then apply Blackwell's theorem on decision-making superiority to the problem of data valuation in machine learning. Consider the true underlying model parameter $\bth$ as the state of the world and the dataset $D$ as a signal about $\bth$. Suppose we aim to use $D$ to select a hypothesis or trained model $h$ from a hypothesis/model class $\mathcal{H}$, which serves as the action space. The utility function $u( h, \bth)$ represents the negative expected loss when the true model parameter is $\bth$ and the hypothesis/model $h$ is chosen:
\[
u(h, \bth) = -\mathbb{E}_{\x,y \sim p(\x,y|\bth)}[l(h(\x), y)] \triangleq -L( h, \bth),
\]
where $l(\cdot)$ is a loss function.

Now, suppose we have a data curation strategy $f(D)$ that reduces the informativeness of the dataset $D$ about $\bth$ in the Blackwell order, i.e., $\bth \to D \to f(D)$ forms a Markov chain. By Blackwell's theorem on decision-making superiority, the decision-maker can achieve at least as low an expected loss using the original dataset $D$ as they can using the curated dataset $f(D)$.

\begin{theorem}\label{thm:blackwell_decision}
    Let $\bth$ be the true underlying model parameter, $D_1$ be a dataset consisting of data points $(\x,y)$ drawn from $p(\x,y|\bth)$, and $D_2$ be a less informative dataset such that $\bth \to D_1 \to D_2$ forms a Markov chain. Consider the decision problem of selecting a hypothesis/trained model $h$ from a hypothesis/model class $\mathcal{H}$ to minimize the expected loss $\mathbb{E}[l(h(\x), y)]$ using a dataset. Then, the minimum expected loss achievable using $D_1$ is at least as low as that achievable using $D_2$. Formally:
    \[
    \min_{h_1: \mathcal{D} \to \mathcal{H}} \mathbb{E}[L(h_1(D_1), \bth)] \leq \min_{h_2: \mathcal{D} \to \mathcal{H}} \mathbb{E}[L( h_2(D_2), \bth)],
    \]
    where $L(h, \bth) = \mathbb{E}_{\x,y \sim p(\x,y|\bth)}[l(h(\x), y)]$ represents the expected loss when the true parameter is $\bth$ and the model $h$ is chosen. The expectation is taken over $\bth \sim p(\bth)$, $D_1$, and $D_2$.
\end{theorem}

\section{Mutual information and proper scoring rules} \label{app:scoring_rules}

Due to the data processing inequality, the simplest metric that might yield a strategy-proof scoring function is the Shannon mutual information  of the model parameter $\bth$ and the curated dataset $f(D) =\widehat D$, denoted by $I(\bth, f(D))$.

One might ask whether it is possible to design a strategy-proof scoring function by estimating the mutual information \( I(\bth, f(D)) \) using only  \( \hat{D}_1, \dots, \hat{D}_k \), bypassing the need for test data \( T \). A natural approach would be to compute
\begin{align}
\label{eqn:mi_theta}
\frac{1}{k} \sum_{i=1}^k \int_\theta p(\bth | \hat{D}_i) \log \frac{p(\bth | \hat{D}_i)}{p(\bth)} \, d\bth
\end{align}
by Monte Carlo integration, i.e., by sampling $\tilde \bth$ from $p(\bth | \hat{D}_i)$ and computing the average
\[
\log \frac{p(\bth = \tilde \bth \mid \hat{D}_i)}{p(\bth = \tilde \bth)}.
\]

The answer is negative. Consider the following strategic data curation method that generates a fake dataset: it converts any dataset $D$ into a fixed dataset $$x^* = \arg\max_{x'} \int_\theta p(\theta|D=x') \log \frac{p(\theta|D=x')}{p(\theta)} d\theta,$$ which represents the optimal dataset $x^*$ that maximizes the score in~\Cref{eqn:mi_theta}. This strategic method would maximize the  score computed by \Cref{eqn:mi_theta}, so \Cref{eqn:mi_theta} does not give a strategy-proof scoring function.

To understand why \Cref{eqn:mi_theta} fails and how it differs from our PMI dataset score, we  introduce proper scoring rules and their connection to our problem.

\paragraph{Proper scoring rules.}
Consider the problem of designing \emph{proper scoring rules} for probabilistic forecasts. Let \( Q \) be a forecast (represented as a probability distribution) and \( S(Q, y) \) be a scoring rule that evaluates \( Q \) based on the observed true outcome \( y \). Suppose the true distribution of \( y \) is \( P \). A scoring rule is called \emph{proper} if it incentivizes the forecaster to report the true distribution \( P \), meaning:

\[
\mathbb{E}_{y \sim P}[S(P, y)] \geq \mathbb{E}_{y \sim P}[S(Q, y)] \quad \text{for all } Q.
\]
Equality holds if and only if \( Q = P \). A well-known example of a proper scoring rule is the \emph{logarithmic scoring rule} \( S(Q, y) = \log Q(y) \), as
\[
\mathbb{E}_{y \sim P}[\log P(y)] - \mathbb{E}_{y \sim P}[\log Q(y)] = D_{KL}(P\Vert Q)\ge 0, \quad \text{for all } Q,
\]
where $D_{KL}(\cdot)$ is the KL-divergence.

However, if now we replace the true observed outcome $y$ with a ``fake'' outcome $\tilde y \sim Q(y)$, the log scoring rule $S(Q, y) = \log Q(\tilde y)$ will no longer be proper. As we will have the expected score
\[
\mathbb{E}[S(Q, y)] =\mathbb{E}\log Q(\tilde y) = \mathbb{E}_{\tilde y \sim Q}\log Q(\tilde y) = -H(Q),
\]
which is the negative entropy of $Q$ and will always be maximized by a deterministic distribution.

\paragraph{Connection between scoring rules and our problem}
Our problem of evaluating data curation strategies can be viewed as a scoring rule design problem through the following mapping:
\begin{itemize}
	\item A dataset $x$ induces a posterior distribution over $\bth$, which we treat as a forecast: $Q_x(\bth) \equiv p(\bth|D=x)$.
	\item A data curation strategy $f(\cdot)$ modifies this forecast to produce $Q_{f(x)}(\bth) = p(\bth|D=f(x))$.
\end{itemize}

So our problem can be viewed as evaluating forecasts about $\bth$, i.e., evaluating $Q_{f(x)}(\bth)=p(\bth|D=f(x))$. Then computing \Cref{eqn:mi_theta} by Monte Carlo integration will be equivalent to first drawing a ``fake'' outcome $\tilde \bth \sim p(\bth|D = f(x))$, and then evaluating $p(\bth|D = f(x))$ by the fake outcome using $ \log\frac{p(\bth = \tilde \bth|D = f(x))}{p(\bth = \tilde \bth)}$. Then  the expected score will always be maximized by setting  $f(x) \equiv x^* = \arg \max_{x'} \int p( \bth|D = x') \log \frac{p(\bth|D = x')}{p( \bth)} \, d\bth$. This mirrors why $S(Q,y) = \log Q(\tilde y)$ fails to be a proper scoring rule.

In contrast, our PMI dataset score can be viewed as a proper scoring rule. Let $p(\bth, D,T)$ be the joint distribution of the true parameter and datasets. Consider the identity function $\text{id}(D) \equiv D$ and the strategic data curation method $f^*(D) \equiv x^* = \arg\max_{x'} \int_\bth p(\bth|D=x') \log \frac{p(\bth|D=x')}{p(\bth)} d\bth$ (here $\bth$ can be replaced by $T$). By our PMI scoring function, the expected score received by $f^*(\cdot)$ is
\begin{align*}
S(f^*(\cdot)) = \E_{T} \left[\log \frac{p(T|D=x^*)}{p(T)}\right] = \E_{T} [\log p(T|D=x^*)] - \E_T \log p(T)
\end{align*}
 and the expected score received by $\text{id}(\cdot)$ is
 \begin{align*}
S(\text{id}(\cdot)) = \E_{D,T} \left[\log \frac{p(T|D)}{p(T)}\right] = \E_{D} \,\E_{T\sim p(T|D)} \left[\log p(T|D)\right] - \E_T \log p(T).
\end{align*}
Then we have
\begin{align*}
	S(\text{id}(\cdot)) - S(f^*(\cdot)) & = \E_{D} \,\E_{T\sim p(T|D)} \left[\log p(T|D)-\log p(T|D=x^*)\right]\\
	& = \E_D \, D_{\text KL}(p(T|D)\Vert p(T|D=x^*))\\
	& \ge 0.
\end{align*}
Therefore, our method properly assigns a lower score to this strategic method.

\section{Integral PMI score} \label{app:integral}
\citet{kong2018water} proposed a method to compute the PMI.
\begin{theorem}[Integral PMI score \citep{kong2018water}] \label{thm:pmi_integral}
The pointwise mutual information
$
    PMI(d, t) = \log \int_\bth p(\bth|D = d)p(\bth|T = t)/p(\bth) \, d\bth.
$ Therefore the data valuation function $U(d,t) = \log \int_\bth p(\bth|D = d)p(\bth|T = t)/p(\bth) \, d\bth$ is truthful.
\end{theorem}
Nonetheless, this integral formulation remains computationally challenging for many basic Bayesian machine learning scenarios. \citet{chen2020truthful} introduced a theoretical framework for evaluating the integral score specifically within exponential family distributions; however, applying their approach is non-trivial. Computing their normalization function $g(\cdot)$ may necessitate solving a non-trivial integral.

For completeness, we provide a stand-alone proof for~\Cref{thm:pmi_integral}.
\begin{theorem}[\cite{kong2018water,chen2020truthful}] \label{lem:MI}
Let $D$ and $T$ be two datasets that are independent conditional on $\bth$, i.e.,
$$
p(D,T|\bth)=p(D|\bth)p(T|\bth),
$$
then the valuation function
\begin{align*}
    U(d, t) = \log \int_\bth p(\bth|D = d)p(\bth|T = t)/p(\bth) \, d\bth.
\end{align*}
is truthful.

\end{theorem}
\begin{proof}
This is basically because when $D$ and $T$ are conditionally independent, we have
\begin{align*}
	U(d', t) &= \log \int_\bth \frac{p(\bth|D = d')p(\bth|T = t)}{p(\bth)} \, d\bth \\
	& = \log \int_\bth \frac{p(d'|\bth)p( t|\bth)p(\bth)}{p(d')p( t)} \, d\bth\\
	& = \log  \frac{\int_\bth p(d', t,\bth) \, d\bth}{p(d')p(t)} \\
	& = \log  \frac{ p(d', t) }{p(d')p(t)} \\
	&= \log \frac{p(t|D=d')}{p(t)}\\
	& =\log p(t|D=d') - \log p(t),
\end{align*}
which is just the log scoring rule. If the data provider manipulates the dataset and reports $f(d)= d' \neq d$, then we have
\begin{align*}
& \E_{T} [U(d, T)|D = d] - \E_{T} [U(d', T)|D = d]	\\
= & \sum_{t\in \mathcal{T}} p(t|D=d) \log p(t|D=d) - \sum_{t\in \mathcal{T}} p(t|D=d) \log p(t|D=d')\\
= & \sum_{t\in \mathcal{T}} p(t|D=d) \log \frac{p(t|D=d)}{p(t|D=d')}\\
= & \ D_{KL}\big(p(t|D=d), p(t|D=d')\big) \\
\ge & \ 0.
\end{align*}

\end{proof}

\section[Missing proofs in the framework section]{Missing proofs in \Cref{sec:framework}}

\subsection[Proof of the mutual-information metric result]{Proof of \Cref{prop:MI_metric}} \label{app:MI_metric}

	Let $D$ and $T$ be two datasets induced by the data generating process described in \Cref{sec:model}, and let $f(D)$ be any strategic data curation method so that $\bth \to D \to f(D)$ forms a Markov chain. We want to show that the Shannon mutual information $I(f(D),T)$, if computable, is a desirable scoring function, in other words, $I(f(D),T) \le I(D,T)$. Due to the data processing inequality, it suffices to prove that $T \to D \to f(D)$ forms a Markov chain.

Since $\bth \to D \to f(D)$ forms a Markov chain, which means that $\bth$ and $f(D)$ are independent conditioned on $D$, and $D$ and $T$ are independent conditioned on $\bth$ by the data generating process, it follows that $T$ and $f(D)$ are independent conditioned on $D$,
\begin{align*}
&p(T, f(D)|D) \\
= & \int_\bth p(T, f(D), \bth|D) \, d\bth\\
= & \int_\bth p(T, f(D)|\bth, D) p(\bth|D) \, d\bth\\
= &\int_\bth p(T| f(D),\bth, D) p(f(D)|\bth, D) p(\bth|D) \, d\bth\\
= &\int_\bth p(T| \bth) p(f(D)| D) p(\bth|D) \, d\bth\\
= & p(f(D)| D) \int_\bth p(T| \bth)  p(\bth|D) \, d\bth\\
= & p(f(D)| D)  p(T|D).\\
\end{align*}
Therefore $T \to D \to f(D)$ forms a Markov chain as well, and by the data processing inequality, the Shannon mutual information of the curated dataset and the test dataset $I(f(D),T)$ will be a desirable scoring function if computable.

\subsection[Proof of the convergence corollary]{Proof of \Cref{cor:convergence}} \label{app:convergence}
We first prove that
\[
\Pr\left(\left|\frac{1}{k} \sum_{i=1}^k U_\eta(\widehat D_i, T_i) - I(\widehat D,T)\right|  \le \varepsilon \right)\ge 1-\delta
\]
when \(k = O\left(\frac{\log(1/\delta)}{\varepsilon^2}\right)\). When the posteriors belong to an exponential family and the datasets have bounded sufficient statistics, there exists a constant \(M\) such that the PMI is bounded, i.e., \(U_\eta(\widehat D_i, T_i) \le M\). Under this condition, the concentration bound follows directly from the Chernoff bound.

The assumption of bounded sufficient statistics is similar to that in \citet{belghazi2021minemutualinformationneural}, who assume that the output of the \emph{statistics network} is bounded (see the assumptions in their Theorem 3). The dependence on \(d\) in their bound,
\(
O\left(\frac{d \log(\sqrt{d}/\varepsilon) + d + \log(1/\delta)}{\varepsilon^2}\right)
\)
mainly comes from a uniform convergence bound, which requires covering a subspace of \(\mathbb{R}^d\) with a finite number of small balls (see their proof of Theorem 6).

When the PMI is not bounded, we can still obtain a result of the same order by applying a clipping procedure. For any positive constant \(M\), define:
\[
f_M(i) = \mathbf{1}\big(U_\eta(\widehat D_i, T_i) > M\big)\cdot U_\eta(\widehat D_i, T_i), \quad
g_M(i) = -\mathbf{1}\big(U_\eta(\widehat D_i, T_i) < -M\big)\cdot U_\eta(\widehat D_i, T_i).
\]
Since \(\mathbb{E}[U_\eta(\widehat D_i, T_i)] = I(\widehat D_i, T_i)\) exists (without loss of generality), the dominated convergence theorem guarantees the existence of a constant \(M\) such that:
\[
\mathbb{E}[f_M(i)] \le \varepsilon/2, \quad \mathbb{E}[g_M(i)] \le \varepsilon/2.
\]
We then clip the PMI at \(\pm M\) and compute:
\begin{align}\label{eqn:clipped}
\frac{1}{k} \sum_{i=1}^k U_\eta(\widehat D_i, T_i) \cdot \mathbf{1}\big(|U_\eta(\widehat D_i, T_i)| \le M\big).
\end{align}
This clipping introduces a bias \(\in [-\varepsilon/2, \varepsilon/2]\). Furthermore, by the Chernoff bound, the clipped estimator in~\Cref{eqn:clipped} converges to its expectation within accuracy \(\varepsilon/2\) with probability at least \(1 - \delta\) when \(k = O\left(\frac{\log(1/\delta)}{\varepsilon^2}\right)\). This completes our argument.

For the expected square error, it is equal to the variance of the estimator since the estimator is unbiased, which decreases as $O(1/k)$.

\subsection{Logistic regression with Gaussian approximation} \label{app:example}

Consider logistic regression with likelihood function $p(y|\x, \bth) = \text{Ber}(y|\text{Sigm}(\bth^T \x))$,
and consider Bayesian logistic regression with Gaussian approximation (see Chapter 8 of~\citet{murphy2012machine}) where a  Gaussian prior $p(\bth) = \mathcal{N}(0, \Sigma_0)$ is assumed. Then given a dataset $(\X, \y)$, (where matrix $\X$ is the input data with each column being a data feature and vector $\y$ is the observed labels,) the approximate posterior is given by $p(\bth|\X,\y) \approx \mathcal{N}(\mu,\Sigma)$ with
\begin{align*}
\mu = \arg\min_\w E(\w), \quad \Sigma^{-1} = \nabla^2 E(\w)|\mu,
\end{align*}
where
$E(\w)$ $= -(\log p(\y|\X, \w)$ $+ \log p(\w)). $
Then $\mu$ can be solved by gradient descent and the Hessian matrix $\Sigma^{-1}$ can be computed in closed form. In particular, if we pick $\Sigma_0 = \mathbf{I}$, then we have $\Sigma^{-1} = \X^T \mathbf{S} \X + \mathbf{I}$,
where $\mathbf{S} = \text{diag}\big(\text{Sigm}(\mu^T \x_i)(1-\text{Sigm}(\mu^T \x_i))\big)$.
Therefore as long as the data collector knows the prior $\mathcal{N}(0, \Sigma_0)$, she will be able to compute the posterior given any dataset, and thus our PMI score can be computed by~\Cref{thm:compute}.
Again, we do not need to assume the distribution of the feature $p(\x|\bth)$ and our PMI score can be used when the test data and the evaluated data have different feature distributions.

\paragraph{Gaussian models.} We provide the closed-form solution for the widely-used Gaussian models below. Consider a Gaussian model with a normal prior $p(\bth) = \mathcal{N}(\mu_0, \Sigma_0)$ and normally distributed posteriors
$
p(\bth|D=d) = \mathcal{N}(\mu_a, \Sigma_a),\ p(\bth|T=t) = \mathcal{N}(\mu_b, \Sigma_b), \ p(\bth|D=d, T=t) = \mathcal{N}(\mu_{ab}, \Sigma_{ab}).
$
We demonstrate that to compute our PMI score, it is sufficient to evaluate just two posteriors: $p(\bth|D=d) = \mathcal{N}(\mu_a, \Sigma_a)$ and $p(\bth|T=t) = \mathcal{N}(\mu_b, \Sigma_b)$.  The parameters of the joint posterior $\mu_{ab}, \Sigma_{ab}$ can be derived from $\mu_a, \Sigma_a, \mu_b, \Sigma_b$.  Consequently, even if data providers are unable to share the entire dataset due to privacy concerns, the PMI score can still be computed as long as the data provider submits $\mu_a$ and $\Sigma_a$.
\begin{corollary}
	\label{thm:compute}
    Suppose we have $p(\bth|D=d) = \mathcal{N}(\mu_a, \Sigma_a)$, $p(\bth|T=t) = \mathcal{N}(\mu_b, \Sigma_b)$, and the prior $p(\bth) = \mathcal{N}(\mu_0, \Sigma_0)$, then our PMI score equals
    \begin{align*}
         U_\eta(d, t) = &\frac{1}{2} \left(\log \frac{\det(\Sigma_0)\det(\tilde{\Sigma})}{\det(\Sigma_a) \det(\Sigma_b)} \right. +  \mu_0^T \Sigma_0 ^{-1} \mu_0   + \tilde{\mu}^T \tilde{\Sigma} ^{-1} \tilde{\mu} - \mu_a^T \Sigma_a ^{-1} \mu_a - \mu_b^T \Sigma_b ^{-1} \mu_b\Bigg),
    \end{align*}
    where $\tilde{\Sigma} = (\Sigma_a^{-1} + \Sigma_b^{-1} - \Sigma_0^{-1})^{-1}$ and $\tilde{\mu} = \tilde{\Sigma} \left( \Sigma_a^{-1} \mu_a + \Sigma_b^{-1} \mu_b - \Sigma_0^{-1} \mu_0 \right)$. In addition, we have $p(\bth|D=d, T=t) = \mathcal{N}(\tilde \mu, \tilde \Sigma)$.
\end{corollary}

\begin{proof}
We consider Gaussian models with posteriors $p(\bth|D=d) = \mathcal{N}(\mu_a, \Sigma_a)$, $p(\bth|T=t) = \mathcal{N}(\mu_b, \Sigma_b)$, $p(\bth|D=d, T=t) = \mathcal{N}(\mu_{ab}, \Sigma_{ab})$, and the prior $p(\bth) = \mathcal{N}(\mu_0, \Sigma_0)$. Then the PMI score with $\eta = 0$ is equal to
\begin{align} \label{eqn:gaussian_eta}
 U_0(d, t) = & \frac{1}{2} \left( \log \frac{\det(\Sigma_0)\det(\Sigma_{ab})}{\det(\Sigma_a) \det(\Sigma_b)} + \mu_0^T \Sigma_0 ^{-1} \mu_0 + \mu_{ab}^T \Sigma_{ab} ^{-1} \mu_{ab} - \mu_a^T \Sigma_a^{-1} \mu_a - \mu_b^T \Sigma_b^{-1} \mu_b \right).
\end{align}
Then it suffices to prove that $\Sigma_{ab} = (\Sigma_a^{-1} + \Sigma_b^{-1} - \Sigma_0^{-1})^{-1}$ and $\mu_{ab} = \Sigma_{ab} \left( \Sigma_a^{-1} \mu_a + \Sigma_b^{-1} \mu_b - \Sigma_0^{-1} \mu_0 \right)$.
Due to conditional independence and according to the proof of \Cref{lem:likelihood_posterior}, we have
\begin{align*}
p(\bth|d,t) & \propto \frac{p(\bth|d) p(\bth|t)}{p(\bth)}\\
 & = \frac{\mathcal{N}(\bth; \mu_a, \Sigma_a)\mathcal{N}(\bth; \mu_b, \Sigma_b)}{\mathcal{N}(\bth; \mu_0, \Sigma_0)}\\
 & \propto \exp\Big( -\frac{1}{2} g(\bth) \Big)
\end{align*}
where
\begin{align*}
    g(\bth) & :=   (\bth- \mu_a) ^T \Sigma_a ^{-1}(\bth - \mu_a) + (\bth- \mu_b) ^T \Sigma_b ^{-1}(\bth - \mu_b)  - (\bth- \mu_0) ^T \Sigma_0 ^{-1}(\bth - \mu_0)
\end{align*}
Here, $g(\bth)$ can be further simplified as $g(\bth) = (\bth -\tilde{\mu}) ^T \tilde{\Sigma}^{-1} (\bth -\tilde{\mu}) +  Z_2$ where
\begin{align*}
    \tilde{\Sigma} &= (\Sigma_a^{-1} + \Sigma_b^{-1} - \Sigma_0^{-1})^{-1}\\
    \tilde{\mu} &= \tilde{\Sigma} \left( \Sigma_a^{-1} \mu_a + \Sigma_b^{-1} \mu_b - \Sigma_0^{-1} \mu_0 \right)\\
    Z_2 &= \mu_a^T \Sigma_a ^{-1} \mu_a + \mu_b^T \Sigma_b ^{-1} \mu_b - \mu_0^T \Sigma_0 ^{-1} \mu_0 - \tilde{\mu}^T \tilde{\Sigma} ^{-1} \tilde{\mu}.
\end{align*}
Then $p(\bth|d,t)$ must be the Gaussian distribution with mean $\tilde{\mu}$ and covariance matrix $\tilde{\Sigma}$.

\end{proof}

\section{Interpretation of PMI}
Our expression in~\Cref{thm:MI} uncovers the relationship between the PMI of two datasets and the predictions they induce about $\bth$.
Using this expression, we demonstrate that the PMI of two datasets can be decomposed into the sum of two terms: (1) a term that measures the similarity between the outcomes obtained from two datasets, i.e., $p(\bth|D)$ and $p(\bth|T)$; (2) a term that measures how much $D, T$ boost the confidence of our estimation of $\bth$.

We first present the interpretation for Gaussian models and then extend it to general distributions.
When the prior $p(\bth)$ is  uninformative compared to $p(\bth|d)$ and $p(\bth|t)$,
the PMI dataset score for Gaussian models can be represented as the sum of two terms: (1) a term quantifying the similarity between $p(\bth|D)$ and $p(\bth|T)$, characterized by the \emph{dual skew G-Jensen-Shannon divergence}~\citep{nielsen2019jensen} between  $p(\bth|D)$ and $p(\bth|T)$; (2) a term assessing how much $D, T$ boost the confidence of our estimation of $\bth$, which is equal to how much $d$ and $t$ reduce the (logarithm of the generalized) variance of our belief about $\bth$.

Given two distributions $p$ and $q$, the dual skew G-Jensen-Shannon divergence between $p$ and $q$ is their total KL divergence to their  geometric mean.
\begin{definition}[Dual skew G-Jensen-Shannon divergence~\citep{nielsen2019jensen}]
The dual skew G-Jensen-Shannon divergence of two distributions $p,q$ for parameter $\alpha\in[0,1]$ is defined as
$ JS_*^{G_\alpha}(p \Vert q) =  (1-\alpha) D_{KL}(G_\alpha(p,q) \Vert p) + \alpha \cdot D_{KL}(G_\alpha(p,q) \Vert q)$, where $G_\alpha(p,q)$ is the weighted geometric mean of $p$ and $q$ with $G_\alpha(p,q)(x)\propto p(x)^{1-\alpha} q(x)^\alpha$.
\end{definition}
Then the PMI dataset score can be expressed as follows.
\begin{theorem}\label{thm:jensen}
	When the prior $p(\bth)$ is  uninformative,  our PMI dataset score for Gaussian models has
	\begin{align*}
	U(d,t) = & \frac{1}{2} \log \frac{|\Sigma_0|}{|\tilde \Sigma|} - 2 \cdot JS_*^{G_\alpha}(\mathcal{N}(\mu_a, \Sigma_a)\Vert \mathcal{N}(\mu_b, \Sigma_b)) - k \log 2
	\end{align*}
	with $\alpha = 1/2$, where $\mathcal{N}(\mu_a, \Sigma_a) = p(\bth|d)$, $\mathcal{N}(\mu_b, \Sigma_b)$ $ = p(\bth|t)$, and $\mathcal{N}(\tilde \mu, \tilde \Sigma)$ $ = p(\bth|d,t)$.
\end{theorem}
The negative dual skew G-Jensen-Shannon divergence indicates the similarity between $p(\bth|D)$ and $p(\bth|T)$. Besides the  constant term $-k\log2$, the term $\frac{1}{2} \log |\Sigma_0|/|\tilde \Sigma|= \frac{1}{2} (\log|\Sigma_0| - \log|\tilde \Sigma|)$ corresponds to the difference in (the logarithm of) the generalized variances of $p(\bth)$ and $p(\bth|d,t)$, as the determinant of the covariance matrix is the generalized variance of a Gaussian distribution. In other words, it could be interpreted as how much $d$ and $t$ reduce the uncertainty or increase the confidence of our estimation. Therefore $\frac{1}{2} \log |\Sigma_0|/|\tilde \Sigma|$ can be interpreted as  how much datasets $d$ and $t$ reduce uncertainty and increase confidence in our estimation.

For general distributions, if we similarly define $D_{\text{KL}}(p(\bth|d,t)\Vert p(\bth|d)) + D_{\text{KL}}(p(\bth|d,t)\Vert p(\bth|t))$ as the divergence and $D_{\text{KL}}(p(\bth|d,t)\Vert p(\bth))$ as the confidence increase, the approximation holds at equality.
See~\Cref{app:jensen} for the proof and the details.
In addition, this KL divergence representation can be interpreted as the  ``mutual information'' of $d$ and $t$ regarding $\bth$.
We discuss this interpretation in~\Cref{app:PMPI}.

\subsection{Interpretation by pointwise mutual parameter information} \label{app:PMPI}
 Firstly,  our score can be represented as $d$ and $t$'s mutual information regarding $\bth$, where the amount of information regarding $\bth$ in a dataset is measured by how much the dataset decreases the KL divergence defined below.

 \begin{definition}[Pointwise parameter information of datasets]
 	Given two datasets $d$, $t$, and a prior $p(\bth)$, define the \emph{pointwise parameter information} of a dataset $s$ as
 	\begin{align*}
 		PI_{d,t}(s) = D_\text{KL}(p(\bth|d,t)\Vert p(\bth)) - D_{\text{KL}}(p(\bth|d,t)\Vert p(\bth|s)),
 	\end{align*}
 	which represents how much observing $s$ reduces the KL divergence to $p(\bth|d,t)$ from our belief about $\bth$. Similarly, we define the \emph{conditional pointwise parameter information} of a dataset $s$ given another dataset $r$ as
 	\begin{align*}
 	  PI_{d,t}(s|r) = & D_\text{KL}(p(\bth|d,t)\Vert p(\bth|r)) - D_{\text{KL}}(p(\bth|d,t)\Vert p(\bth|s,r)),
 	\end{align*}
which represents how much observing $s$ reduces the KL divergence to $p(\bth|d,t)$ if we have already observed $r$.
 \end{definition}
 Then  our score can be represented as  ``mutual information'' similar to the Shannon mutual information $I(X,Y) = H(X) + H(Y) - H(X,Y) = H(X) - H(X|Y)=H(Y)-H(Y|X)$ with the entropy $H(\cdot)$ replaced by our pointwise parameter information.
 \begin{theorem} \label{thm:PMI}
Our PMI score equals
\begin{align}
U_\eta(d,t)  = PI_{d,t}(d) + PI_{d,t}(t) - PI_{d,t}(d \cup t) \notag \triangleq PMI_{d,t}(d,t),
\end{align}
which we define as the pointwise mutual parameter information of $d$ and $t$. In addition, we have
\begin{align*}
	PMI_{d,t}(d,t) = PI_{d,t}(d) - PI_{d,t}(d|t)
	= PI_{d,t}(t) - PI_{d,t}(t|d).
\end{align*}
\end{theorem}
See the proof in~\Cref{app:PMI}. \Cref{thm:PMI} also suggests that our PMI score can be computed by computing/estimating KL divergence between the posteriors.

\subsection[Proof of the PMI theorem]{Proof of \Cref{thm:PMI}} \label{app:PMI}
We prove the theorem by proving the following lemma.
\begin{lemma} \label{lem:KL}
When $D$ and $T$ are independent conditional on $\bth$, we have
$$U_\eta(d,t) = D_\text{KL}(p(\bth|d,t)\Vert p(\bth)) - D_{\text{KL}}(p(\bth|d,t)\Vert p(\bth|d)) - D_{\text{KL}}(p(\bth|d,t)\Vert p(\bth|t)).$$
\end{lemma}
\begin{proof}
	The right side of the equation equals
	\begin{align*}
	&D_\text{KL}(p(\bth|d,t)\Vert p(\bth)) - D_{\text{KL}}(p(\bth|d,t)\Vert p(\bth|d)) - D_{\text{KL}}(p(\bth|d,t)\Vert p(\bth|t))\\
	= & \int p(\bth|d,t)\log\frac{p(\bth|d,t)}{p(\bth)}\, d\bth - \int p(\bth|d,t)\log\frac{p(\bth|d,t)}{p(\bth|d)}\, d\bth - \int p(\bth|d,t)\log\frac{p(\bth|d,t)}{p(\bth|t)}\, d\bth\\
	= & \int p(\bth|d,t)\log\frac{p(\bth|d)p(\bth|t)}{p(\bth|d,t)p(\bth)}\, d\bth \\
	= & \ \log \frac{p(t|d)}{p(t)}\\
	= & \ U_\eta(d,t).
	\end{align*}
The third equation is due to \Cref{thm:MI}, that is, we have $\frac{p(\bth|d)p(\bth|t)}{p(\bth|d,t)p(\bth)}= \frac{p(t|d)}{p(t)}$ for all $\bth$.
\end{proof}
Then according to our definition of pointwise parameter information, we have
\begin{align*}
	U_\eta(d,t)& = D_\text{KL}(p(\bth|d,t)\Vert p(\bth)) - D_{\text{KL}}(p(\bth|d,t)\Vert p(\bth|d)) - D_{\text{KL}}(p(\bth|d,t)\Vert p(\bth|t))\\
	& = \big(D_\text{KL}(p(\bth|d,t)\Vert p(\bth)) - D_{\text{KL}}(p(\bth|d,t)\Vert p(\bth|d))\big) \\
	& \qquad +\big(D_\text{KL}(p(\bth|d,t)\Vert p(\bth)) - D_{\text{KL}}(p(\bth|d,t)\Vert p(\bth|t))\big)\\
	&\qquad - \big( D_{\text{KL}}(p(\bth|d,t)\Vert p(\bth)) - D_{\text{KL}}(p(\bth|d,t)\Vert p(\bth|d,t))\big)\\
	& = PI_{d,t}(d) + PI_{d,t}(t) - PI_{d,t}(d \cup t) \notag\\
&\triangleq PMI_{d,t}(d,t).
\end{align*}
And by our definition of conditional pointwise parameter information, we have
\begin{align*}
	U_\eta(d,t) & = \big(D_\text{KL}(p(\bth|d,t)\Vert p(\bth)) - D_{\text{KL}}(p(\bth|d,t)\Vert p(\bth|d)) \big) \\
	&\qquad - \big( D_{\text{KL}}(p(\bth|d,t)\Vert p(\bth|t)) - D_{\text{KL}}(p(\bth|d,t)\Vert p(\bth|d,t))\big)\\
	& = PI_{d,t}(d) - PI_{d,t}(d|t).
\end{align*}
Similarly, we have $U_\eta(d,t) = PI_{d,t}(t) - PI_{d,t}(t|d)$.

\subsection[Proof of Jensen theorem]{Proof of \Cref{thm:jensen}}\label{app:jensen}
Recall that the dual skew G-Jensen-Shannon divergence is defined as follows.
\begin{definition}[Dual skew G-Jensen-Shannon divergence~\citep{nielsen2019jensen}]
The dual skew G-Jensen-Shannon divergence of two distributions $p,q$ for parameter $\alpha\in[0,1]$ is defined as
$ JS_*^{G_\alpha}(p \Vert q) =  (1-\alpha) D_{KL}(G_\alpha(p,q) \Vert p) + \alpha \cdot D_{KL}(G_\alpha(p,q) \Vert q)$, where $G_\alpha(p,q)$ is the weighted geometric mean of $p$ and $q$ with $G_\alpha(p,q)(x)\propto p(x)^{1-\alpha} q(x)^\alpha$.
\end{definition}
\citet{nielsen2019jensen} solved the dual skew G-Jensen-Shannon divergence JS$^G_*$ between two multivariate Gaussians, which is equal to the following.
\begin{lemma}[\citet{nielsen2019jensen}, Corollary 1]
	The dual skew G-Jensen-Shannon divergence JS$^{G_\alpha}_*$ between two multivariate Gaussians $\mathcal{N}(\mu_1, \Sigma_1)$ and $\mathcal{N}(\mu_2, \Sigma_2)$ with $\alpha = \frac{1}{2}$ is equal to
	\begin{align*}
	JS_*^{G_\alpha}(\mathcal{N}(\mu_1, \Sigma_1)\Vert \mathcal{N}(\mu_2, \Sigma_2)) = &\frac{1}{4}\left( \mu_1^T \Sigma_1^{-1} \mu_1 + \mu_2^T \Sigma_2^{-1} \mu_2 - 2\mu^T \Sigma^{-1}\mu + \log \frac{|\Sigma_1||\Sigma_2|}{|\Sigma|^2} \right),
	\end{align*}
	where $\Sigma = 2(\Sigma_1^{-1} + \Sigma_2^{-1})^{-1}$ and $\mu= \frac{1}{2} \Sigma(\Sigma_1^{-1} \mu_1 + \Sigma_2^{-1} \mu_2)$.
\end{lemma}

Then suppose we have $p(\bth|D=d) = \mathcal{N}(\mu_a, \Sigma_a)$, $p(\bth|T=t) = \mathcal{N}(\mu_b, \Sigma_b)$, the prior $p(\bth) = \mathcal{N}(\mu_0, \Sigma_0)$, and $p(\bth|D=d, T=t) = \mathcal{N}(\tilde \mu, \tilde \Sigma)$ with $\tilde{\Sigma} = (\Sigma_a^{-1} + \Sigma_b^{-1} - \Sigma_0^{-1})^{-1}\approx (\Sigma_a^{-1} + \Sigma_b^{-1})^{-1}$ and $\tilde{\mu} = \tilde{\Sigma} \left( \Sigma_a^{-1} \mu_a + \Sigma_b^{-1} \mu_b - \Sigma_0^{-1} \mu_0 \right) \approx \tilde{\Sigma} \left( \Sigma_a^{-1} \mu_a + \Sigma_b^{-1} \mu_b \right)$.
By definition, we have
\begin{align*}
	JS_*^{G_\alpha}(\mathcal{N}(\mu_a, \Sigma_a)\Vert \mathcal{N}(\mu_b, \Sigma_b))
	= &\frac{1}{4}\left( \mu_a^T \Sigma_a^{-1} \mu_a + \mu_b^T \Sigma_b^{-1} \mu_b - 2\mu^T \Sigma^{-1}\mu + \log \frac{|\Sigma_a||\Sigma_b|}{|\Sigma|^2} \right),
	\end{align*}
	where $\Sigma = 2(\Sigma_a^{-1} + \Sigma_b^{-1})^{-1} \approx 2\tilde \Sigma$ and $\mu= \frac{1}{2} \Sigma(\Sigma_a^{-1} \mu_a + \Sigma_b^{-1} \mu_b) \approx \tilde \mu$.
	Then $U_\eta(d,t)$ defined in~\Cref{thm:compute} has
	\begin{align*}
	U_\eta(d,t) + 2\cdot JS_*^{G_\alpha}(\mathcal{N}(\mu_a, \Sigma_a)\Vert \mathcal{N}(\mu_b, \Sigma_b)) \approx &\frac{1}{2} \log \frac{\det(\Sigma_0)\det(\tilde{\Sigma})}{\det(\Sigma_a) \det(\Sigma_b)} + \frac{1}{2}\log \frac{\det(\Sigma_a)\det(\Sigma_b)}{\det(2 \tilde \Sigma)^2} \\
	=&\frac{1}{2} \log \frac{\det(\Sigma_0)\det(\tilde{\Sigma})}{\det(2 \tilde \Sigma)^2}\\
	=&\frac{1}{2} \log \frac{\det(\Sigma_0)\det(\tilde{\Sigma})}{4^k \cdot \det(\tilde \Sigma)^2}\\
	=&\frac{1}{2} \log \frac{\det(\Sigma_0)}{\det(\tilde \Sigma)} - k \log 2.
	\end{align*}

For general distributions, we can get a similar interpretation using~\Cref{lem:KL}.
Similar to the definition of the dual skew G-Jensen-Shannon divergence, we define $\frac{1}{2}D_{\text{KL}}(p(\bth|d,t)\Vert p(\bth|d)) + \frac{1}{2}D_{\text{KL}}(p(\bth|d,t)\Vert p(\bth|t))$ as the divergence of $p(\bth|d)$ and $p(\bth|t)$, where $p(\bth|d,t)$ is viewed as the geometric mean of $p(\bth|d)$ and $p(\bth|t)$. In addition, we define $D_\text{KL}(p(\bth|d,t)\Vert p(\bth))$ as the counterpart of $\frac{1}{2} \log \frac{\det(\Sigma_0)}{\det(\tilde \Sigma)} - k \log 2$, representing confidence increase/uncertainty reduction. Then by~\Cref{lem:KL}, the PMI dataset score $U_\eta(d,t)$ equals the confidence increase $D_\text{KL}(p(\bth|d,t)\Vert p(\bth))$ minus the divergence $D_{\text{KL}}(p(\bth|d,t)\Vert p(\bth|d)) +D_{\text{KL}}(p(\bth|d,t)\Vert p(\bth|t))$.

\section{Simulations} \label{app:simulation}

\subsection[Detailed experiment setup for MI estimation]{Detailed experiment setup in \Cref{sec:exp_accuracy}} \label{app:exp_accuracy}

\paragraph{A benchmark for dataset MI estimation.}
 Our benchmark generates dataset pairs with pre-defined MI values by resampling from standard datasets such as MNIST, CIFAR, and ImageNet. We construct datasets with binary labels (e.g., label 0 and 1) as follows. Given a target MI value $\lambda$, we first define a joint distribution $p(r_D, r_T)$ of two correlated numbers $r_D, r_T \in \{0.1, 0.2, \dots, 0.9\}$ such that $I(r_D, r_T) = \lambda$. Here, $r_D$ and $r_T$ represent the proportions of 0-labeled images in datasets $D$ and $T$, respectively. Next, we sample $r_D, r_T\sim p(r_D,r_T)$ and sample images to enforce $H(r_D|D) = 0$ and $H(r_T|T) = 0$. This ensures that the ground-truth mutual information $I(D, T) = I(r_D, r_T)$. In our experiments, we generate dataset pairs with ground-truth $I(D, T)\in \{0, 0.1, \dots, 0.9, 1\}$ and assess estimation accuracy by the rank correlation between the estimated and true MI rankings.

In our setting, to establish joint probability distribution $p(r_D, r_T)$ with specified mutual information $\lambda$, we first randomly sample $a_D, a_T \in \{0.1, 0.2,0.3, 0.4\}$ and $b_D, b_T \in \{0.6, 0.7,0.8, 0.9\}$ independently. Then $p(r_D, r_T)$ is defined according to the structure shown in the following table, where the parameter $\rho$ is adjusted to ensure that the mutual information $I(r_D; r_T) = \lambda$.

\begin{center}
\begin{tabular}{|c|c|c|}
 \hline $p(r_D, r_T)$ & $r_D=a_D$ & $r_D=b_D$\\
\hline $r_T = a_T$ & $\rho$ & $\frac{1}{2} - \rho$ \\
 \hline $r_T = b_T$ & $\frac{1}{2} - \rho$ & $\rho$ \\
 \hline
\end{tabular}
\end{center}

We then construct dataset pairs containing images of two classes from MNIST, CIFAR and ImageNet. First, we randomly sample  \(r_D, r_T \sim p(r_D, r_T)\), which represent the proportion of data points with the label \(0\) in datasets \(D\) and \(T\), respectively. Next, we generate a random vector \(L_D \in \{0, 1\}^{N_D}\), where each element is \(0\) with probability \(r_D\), and a similar vector \(L_T \in \{0, 1\}^{N_T}\), where each element is \(0\) with probability \(r_T\). Each value in \(L_D\) and \(L_T\) corresponds to a label for an image. To enforce $H(r_D|D) = 0$ and $H(r_T|T) = 0$, we make a minor modification: we replace the last bit of $L_D$ by $\oplus_{j=1}^{N_D-1} L_D(i) \oplus\mathbf{1}(r_D < 0.5)$ so that the XOR sum of $L_D$ reveals the value of $r_D$. Similarly, we adjust the last bit of $L_T$ so that the XOR sum of $L_T$ reveals $r_T$. Finally, for each label in \(L_D\) and \(L_T\), we pair it with a randomly selected image matching the label, resulting in two correlated datasets \(D\) and \(T\)  with the labels included.

\begin{proposition}
	The mutual information of the generated datasets $I(D,T)=I(r_D,r_T)$.
\end{proposition}
\begin{proof}
	By Theorem 4 of~\citet{gowri2024approximating}, $I(D,T) = I(L_D, L_T)$ assuming that $H(L_D|D)=0$ and $H(L_T|T)=0$. Again, since $L_D$ and $L_T$ fully reveal $r_D$ and $r_T$, we have $H(r_D|L_D) = 0$ and $H(r_T|L_T)=0$, Theorem 4 of~\citet{gowri2024approximating} implies that $I(L_D, L_T) = I(r_D, r_T)$. Therefore $I(D,T)=I(L_D, L_T) = I(r_D,r_T)$.
\end{proof}
To estimate the mutual information $I(D, T) \in \{0, 0.1, \dots, 0.9, 1\}$, we generate $k$ dataset pairs $(D_1, T_1), \dots, (D_k, T_k)$ and compute the average PMI using our formula~\Cref{thm:MI} or other baseline methods.

\paragraph{Methods.}
To compute PMI, we employed the Bayesian logistic regression with Gaussian approximation outlined in \Cref{app:example}.  We use the \texttt{LogisticRegression} function from the \texttt{sklearn} library. The model utilizes the \( L_2 \) norm as the regularization term, with the regularization strength controlled by \( C \). The logistic regression model is configured with a maximum number of iterations set to 5000 (\(\text{max\_iter} = 5000\)) and no intercept fitting (\(\text{fit\_intercept} = \text{False}\)), while all other parameters are set to their default values. The range of \( C \) is tuned via cross-validation.
We computed the estimated mutual information using our PMI formula, averaged over $k=2000$ pairs of datasets with sizes ranging from 50 to 100. The accuracy of our approach is assessed by the Spearman's rank correlation (\(\rho\)) between the estimated mutual information rankings and the true rankings of \( I(D,T) \). For MNIST datasets, we constructed datasets by selecting images labeled as digits "0" and "1". To prevent the dimensionality of the data from being too high, we first applied Principal Component Analysis (PCA) to reduce the dimensionality to 100. For CIFAR datasets, we choose data labeled as "airplane" and "automobile". To capture image features, we first extract image embeddings using ResNet18 pre-trained on ImageNet (with the last layer removed). For ImageNet datasets, we choose data labeled as "cat" and "dog". To capture image features, we first extract image embeddings using ResNet50 pre-trained on ImageNet (with the last layer removed).

We consider the following baselines:
\begin{itemize}[leftmargin=*]

\item \textbf{MINE}: For unstructured high-dimensional data such as images, MINE~\citep{belghazi2021minemutualinformationneural} is one of the most widely used and effective estimators for mutual information~\citep{gowri2024approximating, lee2024benchmarksuiteevaluatingneural}. MINE has been shown to be particularly robust for image-based data, making it a strong baseline in our experiments. In our implementation, we used a three-layer neural network to estimate the mutual information, with the hidden layers having sizes of 1024 units each. The activation function used for all layers was the default ReLU function. The input to the network was formed by concatenating the raw images from the training dataset with their corresponding labels.

\item \textbf{MINE on trained model parameters $\w$}: To further reduce dimensionality, we applied MINE not directly on the raw data but on the trained logistic regression model parameters, $\w$, which serve as a lower-dimensional proxy. The network architecture and hyperparameters used for this baseline were identical to those in the MINE implementation, with the primary difference being that the input to the neural network consisted of the parameters of the trained logistic regression model rather than the raw data.

\item \textbf{LMI}: We adopt a recent approach~\citet{gowri2024approximating}, which is a method designed to estimate MI between high-dimensional variables by leveraging low-dimensional latent representations. Traditional MI estimators struggle with high-dimensional data due to the curse of dimensionality, but LMI addresses this by learning compressed representations of the variables using a cross-predictive autoencoder architecture. This architecture trains encoders and decoders to reconstruct each variable from the latent space of the other, preserving dependency structures while reducing dimensionality. Finally, alternative estimators specifically designed for low-dimensional spaces are applied to approximate MI. In our implementation, we train a four-layer neural network as encoders and decoders with LeakyReLU as activation function. The dimension of latent space is set to be $128$ and we use MINE to approximate MI with low-dimension compressed representations.

\item \textbf{Monte Carlo integration}: We evaluate a simple Monte Carlo integration approach to compute the pointwise mutual information defined in \Cref{prop:MI_metric}. The integration is computed by sampling 2000 points. First, a logistic regression model is trained to obtain the parameters, and then samples are drawn from both the posterior and prior distributions of these parameters. These samples are used to compute the likelihood of the test data. The Monte Carlo estimate of the likelihood is obtained by averaging the likelihoods computed for each sample. In the code, we use the log-sum-exp trick to stabilize the calculation of the logarithms of sums of exponentials. This trick is applied during the computation to avoid numerical instability by shifting the values to a more stable range before taking the logarithm, which helps in preventing overflow or underflow errors when dealing with very large or small numbers.
\end{itemize}

\begin{table}[t]
    \centering
    \begin{tabular}{l c c}
    \hline
    & \textbf{CMNIST} & \textbf{CIFAR} \\
    \cline{2-3}

    \hline
    PMI (C=1)  & 0.954 {\scriptsize $\pm$ 0.021} & - \\
    PMI (C=100)  & 0.967 {\scriptsize $\pm$ 0.015} & -\\
    PMI (C=1000)  & 0.961  {\scriptsize $\pm$ 0.022} & - \\
    PMI (C=10000) & - & 0.976 {\scriptsize $\pm$ 0.000} \\
    PMI (C=20000) & - & 0.976 {\scriptsize $\pm$ 0.001} \\
    PMI (C=50000) & - & 0.982 {\scriptsize $\pm$ 0.000} \\
    \hline
    \end{tabular}
    \caption{
    Spearman's rank correlation (\(\rho\)) between estimated and ground-truth mutual information rankings for PMI on Colored MNIST and CIFAR. PMI consistently achieves high correlation. We sample datasets with size $100$, \(C\) represents the regularization strength, and \(\pm\) denotes the standard deviation over 20 independent trials.
    }
    \label{tab:detail-result-on-4.1}
\end{table}

For each method, we perform 20 trials to compute the standard deviation of its rank correlation.
We conducted our experiments on an Ubuntu system, utilizing 48GB 4090D GPUs for computations requiring GPU acceleration. \autoref{tab:detail-result-on-4.1}, \autoref{tab:detail-result-on-4.1-size50}, and \autoref{figure} show that our PMI estimator consistently achieves high Spearman’s $\rho$ rank correlation, producing accurate estimates with small variances under different choices of regularization strength $C$ and dataset sizes.

\begin{table}[t]
    \centering
    \begin{tabular}{l c c}
    \hline
    & \textbf{CMNIST} & \textbf{CIFAR} \\
    \cline{2-3}

    \hline
    PMI (C=1)  & 0.962 {\scriptsize $\pm$ 0.029} & - \\
    PMI (C=100)  & 0.956 {\scriptsize $\pm$ 0.024} & -\\
    PMI (C=1000)  & 0.964{\scriptsize $\pm$ 0.018}  & - \\
    PMI (C=10000) & - & 0.978{\scriptsize $\pm$ 0.001} \\
    PMI (C=20000) & - & 0.984{\scriptsize $\pm$ 0.000} \\
    PMI (C=50000) & - & 0.986{\scriptsize $\pm$ 0.000} \\
    \hline
    \end{tabular}
    \caption{
    Spearman's rank correlation (\(\rho\)) between estimated and ground-truth mutual information rankings for PMI on Colored MNIST and CIFAR. PMI achieves high correlation regardless of the regularization strength. We sample datasets with size $50$, \(C\) represents the regularization strength.
    }
    \label{tab:detail-result-on-4.1-size50}
\end{table}

\subsection[Detailed experiment setup for curation evaluation]{Detailed experiment setup in \Cref{sec:exp_scoring_evaluation}} \label{app:exp_curation}

\subsubsection{Colored MNIST} \label{app:CMNIST}

\paragraph{Dataset and setup.} We evaluate a logistic regression model with varying regularization strengths $C$ on a colorized MNIST dataset. The prediction task is binary digit classification with labels $0$ and $1$, and the non-essential feature is the background color. This yields four sample groups:
\begin{itemize}
    \item \texttt{blue-label-0} / \texttt{blue-label-1}: Images with a blue background labeled 0 or 1.
    \item \texttt{green-label-0} / \texttt{green-label-1}: Images with a green background labeled 0 or 1.
\end{itemize}

\paragraph{Model and score computation.} The logistic regression model is implemented using the \texttt{Logistic\-Regression} function from the \texttt{sklearn} library. It employs the \( L_2 \) norm as the regularization term, with the strength of regularization controlled by \( C \). The model is configured with a maximum number of iterations set to 5000 (\(\text{max\_iter} = 5000\)) and no intercept fitting (\(\text{fit\_intercept} = \text{False}\)), while all other parameters are set to their default values. The range of \( C \) is tuned via cross-validation.

For each value of \( C \), we sample 1,000 pairs of training and testing datasets. For each pair, we compute the PMI score and test accuracy before and after applying the curation method. To compute the PMI score, we apply the Bayesian logistic regression with a Gaussian approximation as described in \Cref{app:example}. This procedure is independently repeated 10 times, resulting in 10 groups of mean values, each derived from 1,000 repetitions. From these groups, we calculate the overall mean and the standard deviation across the 10 groups.

\paragraph{Data Filtering.} We introduce label noise by flipping the labels of $10\%$ of the training samples. We then remove these mislabeled samples, retrain the model, and record the changes in PMI score and test accuracy.

\paragraph{Data Removal.} We remove samples from the training set based on the non-essential background-color feature and digit-label groups to match the category ratios of the test set. We then retrain the model and record the changes in PMI score and test accuracy.

\paragraph{Results.} Results are reported in \Cref{tab:mnist-results} and \Cref{tab:mnist-results-1}. The two tables use the same curation methods and differ only in the training and test group sizes described in their captions.

\begin{table}[t]
\centering
\begin{tabular}{ccc}
\toprule
\textbf{C} & \textbf{PMI Score Change} & \textbf{Test Accuracy Change (\%)} \\
\midrule
\multicolumn{3}{c}{\textbf{Data Filtering}} \\
10 & 6.4621 $\pm$ 0.8463 & 0.91 $\pm$ 0.06 \\
20 & 6.7529 $\pm$ 0.9451 & 0.80 $\pm$ 0.04 \\
50  & 8.1367 $\pm$ 1.0031 & 0.78 $\pm$ 0.11 \\
100 & 12.7610 $\pm$ 1.2069 & 0.14 $\pm$ 0.01 \\
200 & 13.5874 $\pm$ 1.0423 & 0.31 $\pm$ 0.02 \\
\midrule
\multicolumn{3}{c}{\textbf{Data Removal}} \\
10 & -5.2692 $\pm$ 0.7825 & 0.31 $\pm$ 0.02 \\
20 & -5.1682 $\pm$ 0.8437 & 0.27 $\pm$ 0.02 \\
50 & -5.5265 $\pm$ 0.9823 & 0.28 $\pm$ 0.03 \\
100 & -6.5826 $\pm$ 1.0437 & 0.34 $\pm$ 0.04 \\
200 & -13.9614 $\pm$ 2.0497 & 0.54 $\pm$ 0.04 \\
\bottomrule
\end{tabular}
\caption{Changes in PMI score and test accuracy on Colored MNIST under two data curation methods and different regularization strengths $C$. The original training set, sampled from a larger dataset, consists of samples from four categories, with sizes of $50$ or $100$ per category, and the test set has sizes of $50, 100, 100, 50$. To introduce noise in the \textbf{Filtering} experiment, $10\%$ of the training labels are flipped, and the flipped data points are removed. The \textbf{Removal} method subsamples the training set based on background-color and digit-label groups to match the test-set category ratios. The experiment was repeated $1,000$ times to compute the mean changes in PMI scores and accuracy and repeated $10$ times to compute final means and variances.}
\label{tab:mnist-results}
\end{table}

\begin{table*}[!ht]
\centering
\begin{tabular}{ccc}
\toprule
\textbf{C} & \textbf{PMI Score Change} & \textbf{Test Accuracy Change (\%)} \\
\midrule
\multicolumn{3}{c}{\textbf{Data Filtering}} \\
10 & 7.8126 $\pm$ 0.9157 & 0.92 $\pm$ 0.08 \\
20 & 7.8239 $\pm$ 1.0087 & 0.74 $\pm$ 0.05 \\
50 & 6.2547 $\pm$ 0.9763 & 1.29 $\pm$ 0.09 \\
100 & 14.5329 $\pm$ 1.3924 & 0.20 $\pm$ 0.02 \\
200 & 11.9261 $\pm$ 1.1762 & 0.09 $\pm$ 0.02 \\
\midrule
\multicolumn{3}{c}{\textbf{Data Removal}} \\
10 & -7.2140 $\pm$ 0.9073 & 0.41 $\pm$ 0.03 \\
20 & -7.5783 $\pm$ 1.1306 & 0.19 $\pm$ 0.01 \\
50 & -6.5111 $\pm$ 1.1430 & 0.21 $\pm$ 0.02 \\
100 & -7.1336 $\pm$ 0.9251 & 0.20 $\pm$ 0.02 \\
200 & -14.1899 $\pm$ 1.7394 & 0.67 $\pm$ 0.03 \\
\bottomrule
\end{tabular}
\caption{Changes in PMI score and test accuracy on Colored MNIST under two data curation methods and different regularization strengths $C$. The original training set, sampled from a larger dataset, consists of samples from four categories, with sizes of $50$ or $150$ per category, and the test set has sizes of $50, 150, 150, 50$. To introduce noise in the \textbf{Filtering} experiment, $10\%$ of the training labels are flipped, and the flipped data points are removed. The \textbf{Removal} method subsamples the training set based on background-color and digit-label groups to match the test-set category ratios. The experiment was repeated $1,000$ times to compute the mean changes in PMI scores and accuracy and repeated $10$ times to compute final means and variances.}
\label{tab:mnist-results-1}
\end{table*}

\subsubsection{Corrupted CIFAR}\label{app:CCIFAR}

\paragraph{Dataset and setup.} We evaluate the PMI score on a corrupted CIFAR-10 dataset~\citep{hendrycks2019benchmarkingneuralnetworkrobustness}. The prediction task is binary image classification with labels $0$ and $1$, and the non-essential feature is the corruption type. This yields four data groups:
\begin{itemize}
    \item \texttt{brightness-label-0} / \texttt{brightness-label-1}: Images with \texttt{brightness} corruption labeled $0$ or $1$.
    \item \texttt{contrast-label-0} / \texttt{contrast-label-1}: Images with \texttt{contrast} corruption labeled $0$ or $1$.
\end{itemize}

To investigate robustness, we sample training sets with a balanced ratio ($1:1:1:1$) and evaluate on skewed test sets with ratios of $1:2:2:1$ or $1:3:3:1$ across these four groups.

\paragraph{Model and score computation.} In each experiment, we extract image embeddings using ResNet18 pre-trained on ImageNet (with the last layer removed) and flip $10\%$ of the labels in the sampled training dataset to introduce noise. Then we train logistic regression models on these embeddings with varying regularization strengths $C$ ranging from $10000$ to $100000$.

The logistic regression model is implemented using the \texttt{LogisticRegression} function from the \texttt{sklearn} library. It employs the \( L_2 \) norm as the regularization term, with the strength of regularization controlled by \( C \). The model is configured with a maximum number of iterations set to 5000 (\(\text{max\_iter} = 5000\)) and no intercept fitting (\(\text{fit\_intercept} = \text{False}\)), while all other parameters are set to their default values. We append a dimension whose entries are all 1 to the embeddings and omit the bias term to integrate the bias into the weight vector. The range of \( C \) is tuned via cross-validation.

For each value of $C$, the experiment is repeated 1,000 times, and we compute the mean changes in PMI score and test accuracy across these 1,000 runs. To compute the PMI score, we employ the Bayesian logistic regression with Gaussian approximation outlined in \Cref{app:example}. This process is independently repeated 10 times, producing 10 groups of mean values (each group based on 1,000 repetitions). From these 10 groups, we calculate the overall mean (averaging across all 10,000 experiments) and the standard deviation (calculated from the 10 groups of mean values).

\paragraph{Data Filtering.} We remove the data points whose labels are flipped, retrain the model, and record the changes in PMI score and test accuracy.

\paragraph{Data Removal.} We remove part of the training dataset based on corruption-type and label groups to match the test-set category ratios. We then retrain the model and record the changes in PMI score and test accuracy.

\paragraph{Results.} Results are reported in \Cref{tab:cifar-1-2} and \Cref{tab:cifar-1-3}. The first table uses the $1:2:2:1$ test-set ratio, and the second table uses the $1:3:3:1$ test-set ratio.

\begin{table}[!ht]
\centering
\begin{tabular}{ccc}
\toprule
\textbf{C} & \textbf{PMI Score Change} & \textbf{Test Accuracy Change (\%)} \\
\midrule
\multicolumn{3}{c}{\textbf{Data Filtering}} \\
10000 & 1.8112$\pm$0.1408 & 7.24$\pm$0.07 \\
20000 & 1.7325$\pm$0.0916 & 7.21$\pm$0.11 \\
30000 & 1.6553$\pm$0.1288 & 7.25$\pm$0.17 \\
50000 & 1.5311$\pm$0.1452 & 7.29$\pm$0.10 \\
100000 & 1.2920$\pm$0.1578 & 7.37$\pm$0.14 \\
\midrule
\multicolumn{3}{c}{\textbf{Data Removal}} \\
10000 & -3.8205$\pm$0.0892 & 0.69$\pm$0.07 \\
20000 & -4.1328$\pm$0.1459 & 0.71$\pm$0.16 \\
30000 & -4.4238$\pm$0.0847 & 0.77$\pm$0.10 \\
50000 & -4.6780$\pm$0.1171 & 0.82$\pm$0.13 \\
100000 & -5.0191$\pm$0.0969 & 0.79$\pm$0.16 \\
\bottomrule
\end{tabular}
\caption{Changes in PMI score and test accuracy on Corrupted CIFAR under two data curation methods and different regularization strengths $C$. Here, $C$ denotes the regularization parameter for L2 regularization in the trained logistic regression models, corresponding to a Gaussian prior $N(0, C\cdot \mathbf{I})$.
The training and the test sets consist of $120-180$ samples.
The experiments were repeated 1,000 times to compute the mean changes in PMI scores and test accuracy, and this process was repeated 10 times to estimate the variances.}
\label{tab:cifar-1-2}
\end{table}

\begin{table*}[!ht]
\centering
\begin{tabular}{ccc}
\toprule
\textbf{C} & \textbf{PMI Score Change} & \textbf{Test Accuracy Change (\%)} \\
\midrule
\multicolumn{3}{c}{\textbf{Data Filtering}} \\
10000 & 2.1175$\pm$0.1916 & 7.38$\pm$0.11 \\
20000 & 1.9854$\pm$0.1680 & 7.29$\pm$0.14 \\
30000 & 1.9894$\pm$0.2097 & 7.36$\pm$0.15 \\
50000 & 1.8297$\pm$0.1332 & 7.26$\pm$0.14 \\
100000 & 1.5816$\pm$0.1717 & 7.31$\pm$0.09 \\
\midrule
\multicolumn{3}{c}{\textbf{Data Removal}} \\
10000 & -6.1685$\pm$0.0608 & 1.83$\pm$0.12 \\
20000 & -6.8668$\pm$0.1173 & 2.00$\pm$0.14 \\
30000 & -7.3100$\pm$0.1357 & 1.85$\pm$0.12 \\
50000 & -7.8402$\pm$0.1086 & 1.86$\pm$0.14 \\
100000 & -8.3621$\pm$0.1293 & 1.92$\pm$0.09 \\
\bottomrule
\end{tabular}
\caption{Changes in PMI score and test accuracy on Corrupted CIFAR under two data curation methods and different regularization strengths $C$. The original training set, sampled from a larger dataset, consists of images from four categories, each with size $30$, and the test set has sizes $20, 60, 60, 20$. To introduce noise, $10\%$ of the training labels are flipped. The \textbf{Filtering} method removes flipped data points. The \textbf{Removal} method subsamples the training set to match the test-set category ratios, resulting in sizes of $10, 30, 30, 10$. The experiment was repeated 1,000 times to compute mean changes in PMI scores and accuracy and repeated 10 times to obtain final means and variances.}
\label{tab:cifar-1-3}
\end{table*}

\subsubsection{Stylized ImageNet}

\paragraph{Dataset and setup.} We evaluate the PMI score on the Stylized ImageNet dataset~\citep{geirhos2022imagenettrainedcnnsbiasedtexture}. The prediction task is binary image classification between ``Cat'' and ``Dog'', and the non-essential feature is the artistic style. This yields four data groups:
\begin{itemize}
    \item \texttt{Style A-Cat} / \texttt{Style A-Dog}: Images of cats or dogs stylized with artistic Style A.
    \item \texttt{Style B-Cat} / \texttt{Style B-Dog}: Images of cats or dogs stylized with artistic Style B.
\end{itemize}

\paragraph{Model and score computation.} We employ a ResNet-50 model pretrained on ImageNet to extract feature embeddings. To integrate the bias term directly into the weight vector, we augment the embeddings by appending an additional dimension with a constant value of 1 and omit the explicit bias term during training. We train a logistic regression model using the L-BFGS optimizer with a maximum of 200 iterations. The regularization strength $C$ is tuned via cross-validation within the range $[10^{-3}, 10^{3}]$. All experiments are conducted on NVIDIA 4090D GPUs.

Unlike the Colored MNIST and Corrupted CIFAR settings above, the Stylized ImageNet experiment also varies the training and testing dataset sizes from 80 to 480 samples. Each experimental condition is repeated 200 times to calculate the mean change in the PMI score and test accuracy. This entire process is repeated for 12 independent groups to compute the final mean and standard deviation.

\paragraph{Data Filtering.} We construct training and test sets with equal sizes and a balanced ratio of $1:1:1:1$ across the four groups (Style A-Cat, Style A-Dog, Style B-Cat, Style B-Dog). We randomly flip the labels of $10\%$ of the training data, remove these mislabeled samples, retrain the model, and record the changes in PMI score and test accuracy.

\paragraph{Data Removal.} We construct training and test sets with equal sizes but differing distributions. The training set is initialized with a balanced ratio of $1:1:1:1$, while the test set has a skewed ratio of $1:3:3:1$. We remove samples from the training set to align its distribution with the test-set ratio, retrain the model, and record the changes in PMI score and test accuracy.

\paragraph{Results.} Results are reported in \Cref{tab:stylized_results} and \Cref{fig:stylized_results_plot}. The table reports the additional dataset-size sweep for this setting.

\begin{table}[ht]
    \centering
    \begin{tabular}{ccccc}
        \toprule
         & \multicolumn{2}{c}{\textbf{Data Filtering}} & \multicolumn{2}{c}{\textbf{Data Removal}} \\
        \cmidrule(lr){2-3} \cmidrule(lr){4-5}
        \textbf{Size} & \textbf{PMI Score Change} & \textbf{Acc. Change} & \textbf{PMI Score Change} & \textbf{Acc. Change} \\
        \midrule
        80  & $1.3226\pm0.0431$ & $0.0429\pm0.0023$ & $-1.1958\pm0.0246$ & $0.0185\pm0.0023$ \\
        160 & $4.6813\pm0.0808$ & $0.0407\pm0.0018$ & $-1.5515\pm0.0588$ & $0.0167\pm0.0012$ \\
        240 & $7.6662\pm0.1563$ & $0.0431\pm0.0015$ & $-1.9131\pm0.0843$ & $0.0145\pm0.0014$ \\
        320 & $13.3335\pm0.1557$ & $0.0280\pm0.0010$ & $-2.2702\pm0.0401$ & $0.0119\pm0.0010$ \\
        400 & $17.1730\pm0.1288$ & $0.0282\pm0.0012$ & $-2.5063\pm0.0528$ & $0.0108\pm0.0009$ \\
        480 & $20.9547\pm0.1498$ & $0.0274\pm0.0009$ & $-2.7518\pm0.1264$ & $0.0100\pm0.0007$ \\
        \bottomrule
    \end{tabular}
    \caption{Changes in PMI score and test accuracy on Stylized ImageNet under two data curation methods across dataset sizes. Accuracy changes are reported as fractions. The results are averaged over 12 independent groups, with each group consisting of 200 runs. The ``Size'' column indicates the number of samples in the training/test sets.}
    \label{tab:stylized_results}
\end{table}

\FloatBarrier
\subsection[Detailed experiment implementation for preference experiments]{Detailed experiment implementation in \Cref{sec:exp_preference}} \label{app:exp_preference}
As illustrated in \Cref{tab:dataset_examples}, the Skywork and SHP examples show the distributional shift used to test whether a metric can predict performance on unseen, linguistically distinct distributions.
\begin{table}[h]
\centering
\small
\resizebox{\linewidth}{!}{
\begin{tabular}{@{} >{\centering\arraybackslash}p{0.12\linewidth} p{0.30\linewidth} p{0.58\linewidth} @{}}
\toprule
\textbf{Dataset} & \multicolumn{1}{c}{\textbf{Prompt Example}} & \multicolumn{1}{c}{\textbf{Response Pair Example}} \\ \midrule
\textbf{Skywork} \newline (Math/Code) &
\textit{``Create a mobile app that displays popular sports results.''} &
\begin{minipage}[t]{0.48\linewidth}
    \raggedright
    \textbf{Chosen (Structured):} Sure! Here is a basic outline of how the app could function:
    \begin{itemize}[leftmargin=*, nosep]
        \item Title: Sports Tracker
        \item Features: Home screen ...
    \end{itemize}
\end{minipage}
\hfill
\begin{minipage}[t]{0.48\linewidth}
    \raggedright
    \textbf{Rejected (Vague):} Developing a complete web application is a multi-step process that requires both front-end and back-end development ...
\end{minipage}
\\ \midrule
\textbf{SHP} \newline (Natural) &
\textit{``Successfully defended PhD yesterday. COVID-positive today. Any advice?''} &
\begin{minipage}[t]{0.48\linewidth}
    \raggedright
    \textbf{Chosen (Helpful):} \textit{First, you need to tell your committee for two reasons: first, so they know they were exposed and can react accordingly and second ...}
\end{minipage}
\hfill
\begin{minipage}[t]{0.48\linewidth}
    \raggedright
    \textbf{Rejected (Trivial):} Rest!!!
\end{minipage}
\\ \bottomrule
\end{tabular}
}
\caption{Comparison of prompt and response pair between the training source (Skywork) and the generalization target (SHP).}
\label{tab:dataset_examples}
\end{table}

\subsubsection{Gradient Representation Details}
\label{app:gradient_details}

\paragraph{Mathematical Formulation.}
For a prompt $x$ and response $r$, we first compute the gradient of the standard negative log-likelihood (NLL) loss, defined as $\mathcal{L}_{NLL}(r|x; \theta) = -\log P(r|x; \theta)$.
Given the prohibitive dimensionality of the parameter space ($|\theta| \approx 1.7$B), we employ \textbf{Rademacher random projection} \citep{press2007numerical, rakhshan2020tensorized}  to obtain a compact representation. Specifically, we construct a projection matrix $\Pi \in \mathbb{R}^{|\theta| \times d}$ where each entry is independently drawn from a \textbf{Rademacher distribution}:
\begin{equation}
    \Pi_{ij} \sim \mathcal{U}(\{-1, +1\})
\end{equation}
This implies that each element of the projection matrix is either $+1$ or $-1$ with equal probability $0.5$. According to the Johnson-Lindenstrauss lemma, this random projection preserves the Euclidean distances and inner products between gradient vectors with high probability, provided the target dimension $d$ is sufficiently large.
In our experiments, we set $d=1024$. The final projected embedding vector $\nabla_{\theta}^{\text{proj}} \mathcal{L}(x,r)$ is computed by projecting the raw gradient and scaling it to maintain the expected norm:
\begin{equation}
    \nabla_{\theta}^{\text{proj}} \mathcal{L}(x,r) = \frac{1}{\sqrt{d}} \Pi^\top \nabla_\theta \mathcal{L}_{NLL}(r|x; \theta)
\end{equation}

\paragraph{Input Template.}
To extract gradients that reflect preference features rather than generic semantics, we process each preference pair $(q, r_w, r_l)$ using a specific instruction template. This template explicitly prompts the model to focus on the ``quality'' and ``helpfulness'' of the response during the forward and backward pass:
\begin{quote}
\textit{``Given a user query and a response, generate an embedding that represents the quality and helpfulness of the response. \newline Query: \{$q$\} \newline Response: \{$r$\}''}
\end{quote}

\end{document}